\documentclass[sigconf]{acmart}
\AtBeginDocument{%
  }

\usepackage{enumitem}
\usepackage[ruled,vlined,linesnumbered]{algorithm2e}

\usepackage[ruled,vlined,linesnumbered]{algorithm2e} 
\usepackage{amsthm}
\usepackage{multirow}
\usepackage{booktabs}
\usepackage{xspace}
\newcommand{\defense}{\textsc{PWaveP}\xspace}
\usepackage[table]{xcolor}
\definecolor{Gray}{gray}{0.9}
\usepackage{pifont}

\setcopyright{acmlicensed}
\copyrightyear{2026}
\acmYear{2026}
\setcopyright{cc}
\setcctype{by}
\acmConference[WWW '26] {Proceedings of the ACM Web Conference 2026}{April 13--17, 2026}{Dubai, United Arab Emirates.}
\acmBooktitle{Proceedings of the ACM Web Conference 2026 (WWW '26), April 13--17, 2026, Dubai, United Arab Emirates}
\acmISBN{979-8-4007-2307-0/2026/04}
\acmDOI{10.1145/3774904.3792351}

\begin{document}

\title[\defense: Purifying Imperceptible Adversarial Perturbations in 3D Point Clouds via Spectral Graph Wavelets]{\defense: Purifying Imperceptible Adversarial Perturbations in 3D Point Clouds via Spectral Graph Wavelets}

\author{Haoran Li}
\affiliation{%
	\institution{Software College, Northeastern University}
	\city{Shenyang}
	\country{China}}
\email{lihaoran@stumail.neu.edu.cn}

\author{Renyang Liu}
\authornotemark[1]
\affiliation{%
	\institution{Institute of Data Science, National University of Singapore}
	\city{Singapore}
	\country{Singapore}
}
\email{ryliu@nus.edu.sg}

\author{Hongjia Liu}
\affiliation{%
	\institution{Software College, Northeastern University}
	\city{Shenyang}
	\country{China}}
\email{liuhongjia@stumail.neu.edu.cn}

\author{Chen Wang}
\affiliation{%
	\institution{Software College, Shenyang University of Technology}
	\city{Shenyang}
	\country{China}}
\email{wangchen@sut.edu.cn}

\author{Long Yin}
\affiliation{%
	\institution{Software College, Northeastern University}
	\city{Shenyang}
	\country{China}}
\email{2110499@stu.neu.edu.cn}

\author{Jian Xu}
\authornote{Renyang Liu and Jian Xu are the corresponding authors.}
\affiliation{%
	\institution{Software College, Northeastern University}
	\city{Shenyang}
	\country{China}}
\email{xuj@mail.neu.edu.cn}

\renewcommand{\shortauthors}{Haoran Li et al.}

\begin{abstract}
	Recent progress in adversarial attacks on 3D point clouds, particularly in achieving spatial imperceptibility and high attack performance, presents significant challenges for defenders. Current defensive approaches remain cumbersome, often requiring invasive model modifications, expensive training procedures or auxiliary data access. To address these threats, in this paper, we propose a plug-and-play and non-invasive defense mechanism in the spectral domain, grounded in a theoretical and empirical analysis of the relationship between imperceptible perturbations and high-frequency spectral components. Building upon these insights, we introduce a novel purification framework, termed \defense, which begins by computing a spectral graph wavelet domain saliency score and local sparsity score for each point. Guided by these values, \defense adopts a hierarchical strategy, it eliminates the most salient points, which are identified as hardly recoverable adversarial outliers. Simultaneously, it applies a spectral filtering process to a broader set of moderately salient points. This process leverages a graph wavelet transform to attenuate high-frequency coefficients associated with the targeted points, thereby effectively suppressing adversarial noise. Extensive evaluations demonstrate that the proposed \defense achieves superior accuracy and robustness compared to existing approaches, advancing the state-of-the-art in 3D point cloud purification. Code and datasets are available at \url{https://github.com/a772316182/pwavep}
\end{abstract}



\maketitle

\section{Introduction}
The proliferation of interactive 3D content has become a cornerstone of the modern Web, powering applications from immersive e-commerce and virtual try-ons to augmented reality platforms and collaborative online environments~\cite{chen2017multi}. Central to these Web-based experiences is the 3D point cloud, a fundamental data representation that captures the geometry of objects through discrete points on their surfaces~\cite{guo2020deep}. These point clouds, typically acquired via LiDAR or 3D scanners, are processed by deep learning models to enable rich user interaction and semantic understanding directly within the browser~\cite{ma2022rethinkingnetworkdesignlocal}. This paper addresses the critical challenge of securing these Web-native 3D applications against adversarial manipulation.

The reliance on deep models for processing 3D data exposes a significant vulnerability that directly threatens the integrity and security of these Web services \cite{liu2023model}. Prior studies have demonstrated that the underlying models can be easily deceived by imperceptible adversarial perturbations \cite{liu2025stba}. Through carefully designed optimization strategies~\cite{10.1093/comjnl/bxad109, gao2023imperceptible}, attackers can subtly shift point positions to craft adversarial point clouds that mislead the model while remaining visually indistinguishable from the original. Recent advances have made these attacks even more potent and stealthy~\cite{yang2024hiding, lou2024hide, Tang_Du_Peng_Wang_Liu_Liu_Tian_2025}. Consequently, defending against such attacks is not merely a machine learning problem but a fundamental challenge for the safe and reliable deployment of trustworthy 3D applications on the Web.

To mitigate such attacks, a variety of defense techniques have been developed. However, many existing approaches suffer from significant limitations in practical deployment. (1) {Invasive and Expensive}: Several methods, including adversarial training \cite{huang2024pointcat}, necessitate modifying or fine-tuning the target model, which may be infeasible in scenarios where the model is well-trained and immutable. Other methods require training separate purification networks, often relying on generative models like diffusion-based frameworks \cite{10.1145/3581783.3612018}. These methods typically demand access to clean training data and incur extra training overhead. (2) {Inflexible and destructive}: Many defenses adopt rigid strategies that detect and remove suspicious points based on heuristic thresholds \cite{li2024pointcvar, yang2019adversarial}. Such coarse removal often compromises the geometric fidelity of the point cloud, especially in the case of imperceptible perturbations where most changes manifest as subtle surface-level deviations rather than conspicuous outliers.

To address these challenges, we propose \defense, a novel defense framework that is both practical and effective. First, \defense operates as a self-contained, training-free pipeline that requires neither modification of the target model nor access to clean or auxiliary datasets. Its plug-and-play design ensures broad compatibility with existing systems. In addition, \defense incorporates an adaptive purification strategy guided by a hybrid spectral-spatial saliency score. Based on this analysis, it applies a hierarchical mechanism that removes only the most severely perturbed and irrecoverable points, while attenuating high-frequency adversarial noise in moderately affected regions. This design enables \defense to effectively suppress adversarial perturbations while preserving the structural integrity of the input point cloud.

Specifically, the design of \defense is grounded in our theoretical and empirical analysis of imperceptible adversarial perturbations in the graph spectral domain. To maintain low perceptual distances quantified by standard metrics such as Chamfer Distance and Earth Mover's Distance, adversaries are incentivized to concentrate perturbation energy within high-frequency spectral components. This insight motivates the adoption of the graph wavelet transform, which provides localized spectral filtering capabilities. By leveraging this transform, \defense accurately identifies and suppresses adversarial noise while preserving essential geometric structures. This theoretical foundation underpins the robustness and fidelity achieved by our approach.

Extensive experimental results demonstrate that \defense consistently outperforms existing defense methods across a wide range of models, datasets, and adversarial attacks. It substantially improves classification accuracy from near 0\% to 97\% on adversarial examples, while maintaining high performance on clean inputs with minimal degradation ($\sim$ 1\%). Furthermore, evaluations based on Chamfer Distance confirm its superior capability in preserving the geometric structure of the original point clouds. These findings collectively validate the robustness and practical effectiveness of the proposed spectral wavelet-based purification framework.
Our major contributions are as follows:
\begin{itemize}[leftmargin=*]
	\item We propose \defense, a novel non-invasive purification framework that leverages the graph wavelet transform and a hybrid saliency score for effective hierarchical defense against adversarial perturbations in 3D point clouds.

	\item We provide a comprehensive theoretical and empirical analysis demonstrating that imperceptible adversarial perturbations predominantly concentrate in the high-frequency components of the graph spectral domain.

	\item Extensive experiments validate that \defense substantially improves classification robustness on adversarial examples while maintaining high accuracy on clean data, confirming its practical effectiveness.
\end{itemize}



\section{Backgrounds}
\subsection{Adversarial Attacks on 3D Point Clouds}
3D point cloud classification is a fundamental task in 3D vision. Pioneering work~\cite{qi2017pointnet} enabled deep learning directly on raw point clouds, powering web applications from immersive e-commerce to augmented reality platforms~\cite{chen2017multi}. 

On the other hand, the adversarial attack on 3D point clouds aims to generate a perturbed point cloud to mislead the classifier. The earliest approaches to implement these perturbations were direct adaptations from the 2D image domain, including variants such as FGSM \cite{liu2019extending} and PGD \cite{xiang2019generating}. Although these approaches achieved high attack success rates, they exhibited poor imperceptibility, often resulting in conspicuous outliers detached from the object's surface. Thus, more recent advancements have focused on 3D geometry. For instance, Normal Attack \cite{tang2022normalattack} incorporates both the gradient and tangent directions at each point to provide smooth regularization. SI-ADV \cite{huang2022shape} defines imperceptibility in terms of shape invariance and applies a reversible coordinate transformation to the input PC to preserve its shape. HiT-ADV \cite{lou2024hide} embeds deformation perturbations into regions with drastic curvature changes and complex surfaces, ensuring that shape variations remain visually imperceptible.

\subsection{Adversarial Defenses on 3D Point Clouds}
To mitigate such attacks, various defense methods have been proposed. In the initial phases, since perturbations often deviated from the original surface, a prominent defense method was Statistical Outlier Removal (SOR) \cite{rusu2008towards, yang2019adversarial}, which computes the average distance between point pairs and identifies adversarial points as outliers. However, as attack methods have evolved, the effectiveness of SOR has diminished \cite{ma2020efficient}. To address this, DUP-Net \cite{zhou2019dup} augments the process with an additional up-sampling network. PointDP and Ada3Diff \cite{sun2022pointdp,10.1145/3581783.3612018} employ diffusion models to purify adversarial point clouds. Other approaches \cite{dong2020self, li2022robust, sun2021adversarially} aim to develop classification models robust to adversarial point clouds by designing resilient training methods or incorporating robust networks.

\begin{figure*}[t]
	\centering
	\includegraphics[width=1\linewidth]{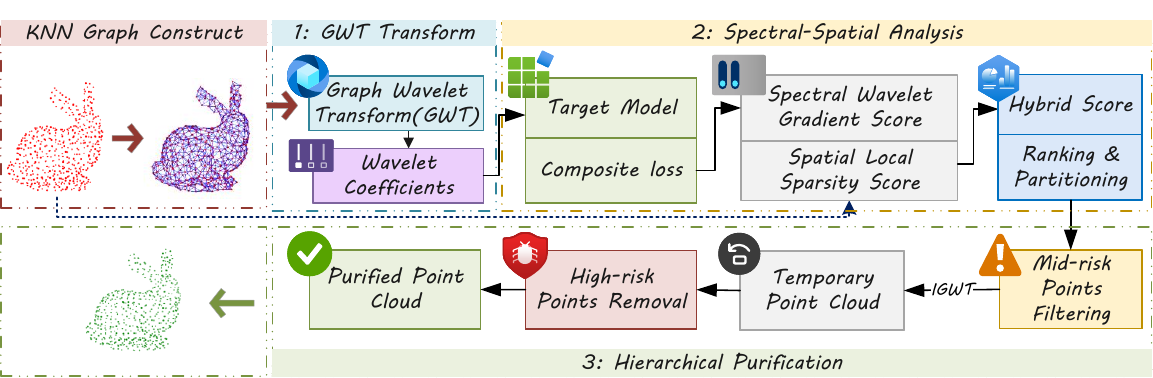}
	\caption{The pipeline of our Point Cloud Wavelet Purification (\defense). (1) The method first applies the Graph Wavelet Transform (GWT) to the PC's KNN graph. (2) A hybrid spectral-spatial analysis then identifies and partitions points into high- and mid-risk sets based on a composite saliency score. (3) Finally, a hierarchical purification module filters mid-risk points and removes high-risk points through an inverse GWT (IGWT) process to restore a clean PC.}
	\label{fig:framework}
\end{figure*}

\subsection{Problem Formulation \& Threat Model}
A 3D Point Cloud (PC) is composed of an unordered set of points $\mathcal{P} = \{p_i \}_{i=1}^N \in \mathbb{R}^{N\times3}$ sampled from the surface of an object with the ground truth label $y_{\mathcal{P}}$, and $N$ is the number of points in the PC. Each point $p_i\in\mathbb{R}^3$ is
a vector representing the 3D coordinates of the point. A classifier $f$ takes the PC $\mathcal{P}$ as the input and outputs
the label prediction $f(\mathcal{P})$.

Given a point cloud $\mathcal{P}$ with real label $y_{\mathcal{P}}$, an adversary generates a perturbed point cloud $\mathcal{P}'$ that misleads the classifier $f$, such that $f(\mathcal{P}') \neq y_{\mathcal{P}}$. The objective is to design a purification operator $d$ that removes adversarial perturbations, yielding a purified point cloud $\mathcal{P}'' = d(\mathcal{P}')$ such that $f(\mathcal{P}'') = f(\mathcal{P}) = y_{\mathcal{P}}$.

In this work, we consider practical deployment scenarios.
For our attacker, we adopt the most threaten setting, namely the white-box attack scenario, where the attacker has full access to both the target model and the dataset. For our defender, we impose two essential constraints: (1) \textit{Non-intrusiveness}, meaning the defender possesses read-only access to the target model; and (2) \textit{Plug-and-play}, meaning the defender cannot access any clean or auxiliary datasets, nor perform any training procedures.

\subsection{Spectral Methods on Point Clouds}
In this paper, our defense primarily operates in the spectral domain of point clouds, we thus introduce some essential concepts related to the our method here. Unlike 2D images, the spectral domain of a point cloud is defined on its $K$-nearest neighbor (KNN) graph. An undirected, unweighted graph with $N$ nodes is represented as $\mathcal{G} = (\mathcal{V}, \mathcal{E})$, where $\mathcal{V}$ and $\mathcal{E}$ denote the node and edge sets, and $A \in \mathbb{R}^{N \times N}$ is the adjacency matrix. The graph spectrum is characterized by the Laplacian matrix $L = D - A$, where $D_{ii} = \sum_j A_{ij}$. As $L$ is real, symmetric, and positive semi-definite, it admits the eigen-decomposition $L = U \Lambda U^\top$, with $U$ containing the eigenvectors and $\Lambda = \mathrm{diag}(\lambda_1, \dots, \lambda_N)$ the diagonal matrix of eigenvalues in non-decreasing order ($0 = \lambda_1 \leq \cdots \leq \lambda_N$). The columns of $U$ form the Graph Fourier Basis, and the eigenvalues represent the graph frequencies~\cite{chung1997spectral}. The smoothness of a signal $h \in \mathbb{R}^N$ on the graph is given by:
\begin{equation}
	h^\top L h = \sum_{(i, j) \in \mathcal{E}} (h_i - h_j)^2 = \sum_{k=1}^N \lambda_k \hat{h}_k^2,
\end{equation}
where $\hat{h} = U^\top h$. It shows that components with smaller eigenvalues correspond to smoother (lower-frequency) signal variations, while larger eigenvalues correspond to higher-frequency variations. The Graph Fourier Transform (GFT) enables spectral manipulation of signals as $\tilde{h} = U \varphi(\Lambda) U^\top h$, where $\varphi$ is an element-wise spectral filter.

By treating the coordinates of a point cloud as three independent signals (one for each axis) on its KNN graph, the structure can be directly analyzed or filtered in the spectral domain~\cite{wen2024pointwavelet, fan2024invisible, liu2024explicitly}. For each suspect point cloud $\mathcal{P}'$, we construct an unweighted, undirected KNN graph $\mathcal{G}$ (empirically $K=20$) to ensure the validity of spectral analysis~\cite{benson2016higher}.

\section{Methodology}
The overview of \defense is illustrated in Figure~\ref{fig:framework}, which comprises three main stages. (1) GWT Wavelets Transform: The input point cloud is represented as a KNN graph and transformed into the spectral domain via the Graph Wavelet Transform (GWT), resulting in wavelet coefficients (Sec.~\ref{subsec:Graph_Wavelets_Transform}). (2) Spectral-Spatial Analysis: We introduce a hybrid saliency metric that combines a Spectral Wavelet Gradient Score—derived from the target model's feedback—and a Spatial Local Sparsity Score that captures geometric anomalies. Points are subsequently ranked and partitioned into high-risk and mid-risk sets (Sec.~\ref{subsec:Spectral-Spatial_Analysis}). (3) Hierarchical Purification: A targeted purification strategy is applied by first attenuating the wavelet coefficients of mid-risk points, reconstructing a temporary point cloud via the inverse GWT (IGWT), and then removing the high-risk points to obtain the final robust point cloud (Sec.~\ref{subsec:Hierarchical_Purification}). Details of each stage are described in the following sections.

\subsection{Properties of Imperceptible Perturbations}
Recently, the frequency perspective has been adopted to investigate the properties of adversarial perturbation \cite{xu2021deep, zhang2024fourier}, the primary advantage of which is its ability to reveal patterns that are otherwise imperceptible in the spatial domain. Although adversarial perturbations are often subtle changes spatially, they can exhibit distinct characteristics in the spectral domain. This provides a unique opportunity to identify and mitigate them. 

To theoretically analyze the properties of the unnoticeable perturbations on PCs so that guide our method design, we first consider the Chamfer Distance (CD), a common metric for measuring discrepancies between two PCs. Note that CD is frequently used as an optimization objective when crafting imperceptible perturbations. Mathematically, CD is defined as:
\begin{equation}
    \mathrm{CD}(\mathcal{P}, \mathcal{P}') = \frac{1}{|\mathcal{P}'|} \sum_{p' \in \mathcal{P}'} \min_{p \in \mathcal{P}} \| p' - p \|_2^2 + \frac{1}{|\mathcal{P}|} \sum_{p \in \mathcal{P}} \min_{p' \in \mathcal{P}'} \| p - p' \|_2^2.
\end{equation}
For the sake of simplicity in our analysis, we assume bijective correspondence of the perturbations $\Delta$ \footnote{We clarify that the bijective correspondence is a simplifying assumption made for the theoretical analysis only.}. That is, $\mathcal{P}'=\mathcal{P}+\Delta$, $p'_i = p_i + \Delta_i$ and $|\mathcal{P}| = |\mathcal{P}'| = N$. Therefore, CD is bounded by:
\begin{equation}
	\mathrm{CD}(\mathcal{P}, \mathcal{P}') \leq \frac{2}{N} \sum_{i=1}^N \|\Delta_i\|_2^2 = \frac{2}{N} \|\Delta\|_F^2,
\end{equation}
where $\Delta$ collects all $\Delta_i$. Let $\hat{\Delta} = U^\top \Delta$ be the spectral-domain perturbations via GFT. By Parseval's theorem, $\|\hat{\Delta}\|_F^2 = \|\Delta\|_F^2$, implying CD constraints limit total spectral energy.
Then, the key question is how an attacker distributes the limited spectral energy to make perturbations unnoticeable. Therefore, we adopt the Earth Mover's Distance (EMD) on the KNN graph, as we already treat the coordinates as signals.

Without loss of generality, we simplify the derivation by analyzing the perturbation on a single coordinate axis. By Kantorovich-Rubinstein duality, EMD equates to maximizing the alignment between a 1-Lipschitz $m$ and the perturbation $\hat{\Delta}=U^\top \Delta$. To make this tractable, we relax the strict Lipschitz constraint to a normalized smoothness constraint based on Dirichlet energy ($m^\top L m \le |\mathcal{E}|$):
\begin{align}
\mathrm{EMD}(\mathcal{P}, \mathcal{P}') \leq \frac{1}{N} \sup_{m: m^\top L m \le |\mathcal{E}|} \langle m, \Delta \rangle.
\end{align}
Assuming $\Delta$ is band-limited with spectral support within $[\lambda_{\min}\neq0, \lambda_{\max}]$, and applying the Cauchy-Schwarz inequality, yields a spectral bound related to the Sobolev norm:
\begin{align}
\mathrm{EMD}(\mathcal{P}, \mathcal{P}') \le \frac{\sqrt{|\mathcal{E}|}}{N} \sqrt{\sum_{k=2}^N \frac{\hat{\Delta}_k^2}{\lambda_k}} \le \frac{C}{\sqrt{\lambda_{\min}}} ||\Delta||_2.
\end{align}
This bound reveals that EMD sensitivity is inversely proportional to the square root of the frequency $\lambda$.
We analyze low/high-frequency components: (i) For low-frequency $\Delta_L$ (small $\lambda$), the weight $1/\sqrt{\lambda}$ is large, amplifying the transport cost as coherent distortions require moving mass over long distances. (ii) For high-frequency $\Delta_H$ (large $\lambda$), the term $1/\sqrt{\lambda}$ heavily penalizes the metric. Spatially, this corresponds to local oscillations where noise cancels out within neighborhoods, resulting in minimal mass transport. Thus, for fixed energy, high-frequency perturbations yield significantly smaller EMD. To remain unnoticeable, attackers favor high-frequency distributions, forming the basis of our purification strategy.
Empirical results on real-world datasets are provided in Sec.~\ref{subsec:main_results}.

\subsection{Graph Wavelets Transform}
\label{subsec:Graph_Wavelets_Transform}
After suggesting that the perturbations are concentrated in the high-frequency region, a straightforward approach is to use the GFT to construct a low-pass filter to eliminate the perturbations, e.g., $\varphi(\lambda_k) = 1 - \frac{\lambda_k}{\lambda_{max}}$.
However, adversarial perturbations on point clouds are often spatially localized, and the global nature of the GFT basis causes a local perturbation's energy to "smear" across the entire spectrum. This makes precise, targeted filtering impossible without distorting clean geometry. We therefore pivot to the Graph Wavelet Transform (GWT), which provides the necessary spatial and spectral localization to address this challenge.

Basically, GWT extends the GFT framework by employing a more sophisticated set of filters designed for multi-scale, localized analysis. Instead of a single low-pass filter $\phi(\cdot)$, GWT introduces other two key components: a wavelet kernel $g(\cdot)$ and a scale parameter $s$. The kernel $g$ is specifically designed as a band-pass filter in the spectral domain ($g(0)=0, g(\infty)\to0$), allowing it to isolate signal variations within specific frequency bands.
Specifically, the wavelet operator at scale $s$ is a spectral filter defined as:
\begin{equation}
	T_s = g_s(L) = U g_s(\Lambda) U^{\top}.
	\label{eq:gwt_operator}
\end{equation}
Applying this operator to a signal $h$ yields the wavelet coefficients $\psi_{s,i}$, which are defined as the output of the filter at each node $i$:
\begin{equation}
	\psi_{s,i} = (T_s h)_i
	\label{eq:gwt_coeffs},
\end{equation}
in which each coefficient $\psi_{s,i}$ provides a spatially localized measurement of the signal $h$'s content around node $i$ at scale $s$. To ensure a complete signal representation, a low-pass scaling function $T_\phi = \phi(L)$ (akin to a "father wavelet") is used alongside the band-pass wavelets to capture the foundational, low-frequency structure of the signal. Similarly, the scaling coefficients are: $\delta_i = (T_\phi h)_i$.

GWT is invertible, allowing the original signal to be  reconstructed from its constituent components. The inverse GWT (IGWT) synthesizes the signal by combining the scaling coefficients $\delta$ with the wavelet coefficients $\psi_s$.
Specifically, for $S$ discrete scales, $h$ can be recovered by:
\begin{equation}
	\tilde{h} = T_\phi \delta + \sum_{s=1}^{S} T_s \psi_s.
	\label{eq:igwt_reconstruction}
\end{equation}
Note that this reconstruction is exact when the scaling and wavelet kernels are designed to form a tight frame; if not, we may need to compute a pseudo-inverse. For a detailed description of GWT, we refer the reader to \cite{HAMMOND2011129,6494675}.

\noindent\textbf{Choice of the Wavelet Kernel}
In GWT, popular kernel implementations include Mexican Hat and Meyer family. In which, Meyer offers excellent orthogonality and smoothness for perfect decomposition and reconstruction. Mexican Hat excels as a feature detector, isolating local anomalies by filtering smooth backgrounds and intense noise. We adopt the Mexican Hat kernel in this work. A experimental analysis of the impact of kernel choice on performance is provided in Sec.~\ref{subsec:ablation_study}.

\noindent\textbf{Chebyshev Approximation} Eigen-decomposition has $O(N^3)$ complexity. Fortunately, we can use Chebyshev expansion \cite{defferrard2016convolutional} to approximate $T_s, T_\phi$; detailed in Appendix \ref{app_sec:Chebyshev Approximation}.

\subsection{Spectral-Spatial Analysis}
\label{subsec:Spectral-Spatial_Analysis}
\subsubsection{Spectral Wavelet Gradient Score}
Given that adversarial perturbations are primarily concentrated in high-frequency regions, we purify the PC by modifying the wavelet coefficients \(\psi_s\) while keeping the scaling coefficients fixed, followed by reconstruction using the IGWT as in Eq.~\ref{eq:igwt_reconstruction}.

However, determining the optimal modifications to \(\psi_s\) is challenging. To address this, we integrate the IGWT into the target model \(f\) to extract saliency:
\begin{equation}
	\frac{\partial \mathcal{L}}{\partial \psi_s}
	= \frac{\partial \mathcal{L}}{\partial f} \cdot \frac{\partial f}{\partial \mathcal{P}'} \cdot \frac{\partial \mathcal{P}'}{\partial \psi_s}
	= \frac{\partial \mathcal{L}}{\partial \mathcal{P}'} \cdot T_s,
\end{equation}
this chain rule  presents a practical approach: without modifying the model, simply projecting the gradient $\frac{\partial \mathcal{L}}{\partial \mathcal{P}'}$ to the wavelet domain enables the desired analysis. Specifically,  \(\mathcal{L}\) is a composite loss function that incorporates both cross-entropy loss and the stability of the model's internal feature representations:
\begin{equation}
	\label{hyper-alpha}
	\mathcal{L} = \mathcal{L}_{CE} + \alpha \|\mathcal{F}_l({\mathcal{P}'})\|_2,
\end{equation}
with \(\mathcal{L}_{CE}\) using the model's predictions as pseudo labels, \(\alpha\) a hyperparameter, and \(\mathcal{F}_l\) the feature map at a pre-selected intermediate layer \(l\). For the impact of \(l\), please refer to Appendix \ref{sec:Intermediate Layer Choice}.

\subsubsection{Spatial Local Sparsity Score}
Since our KNN graph is unweighted and undirected, some geometric information may be overlooked, making reliance on \(\psi_s\) saliency alone insufficient. Therefore, to highlight such information, we construct a hybrid saliency score that fuses model gradients and intrinsic geometric properties. Inspired by \cite{rusu2008towards}, we define the Local Sparsity Score (LSS) for each point \(p_i\) based on Euclidean distances, yielding Spectral-Spatial Hybrid Analysis:
\begin{equation}
	\label{eq:lss}
	S_{LSS}(p_i) = (d_i - \bar{d})^2,
\end{equation}
where $d_i = \frac{1}{K} \sum_{j=1}^N A_{ij} \|p_i - p_j\|_2$ is the average Euclidean distance from point $p_i$ to its neighbors, then $\bar{d} = \frac{1}{N} \sum_{k=1}^N d_k$ is the global average of these local distances across the entire original suspect PC $\mathcal{P}'$.

\subsubsection{Ranking and Partitioning}
We then compute the hybrid saliency by combining high-frequency spectral saliency and geometric score for each $p_i\in\mathcal{P}'$:
\begin{equation}
	S_{\text{hy}}(p_i) = \sum_{s=\frac{S}{2}}^S \|\nabla_{\psi_{s,i}} \mathcal{L}\|_2 + \beta S_{LSS}(p_i),
	\label{eq:hybrid-saliency}
\end{equation}
where \(\beta\) is a balancing hyperparameter.
We then categorize points based on the descending rank of their hybrid saliency score $S_{\text{hy}}$. The high-risk set ($\mathcal{P}_{\text{high}}'$) consists of the top 1\% of points, and the mid-risk set ($\mathcal{P}_{\text{mid}}'$) comprises the next 9\%:
\begin{align}
	\label{eq:high_risk_set_simple}
	\mathcal{P}_{\text{high}}' & = \{p_i \mid \text{Rank}(p_i) \leq 0.01N\};        \\
	\label{eq:mid_risk_set_simple}
	\mathcal{P}_{\text{mid}}'  & = \{p_i \mid 0.01N < \text{Rank}(p_i) \leq 0.1N\},
\end{align}
where $\text{Rank}(p)$ is the rank of point $p$ when all points are sorted by their saliency score $S_{\text{hy}}$ in descending order.

\subsection{Hierarchical Purification}
\label{subsec:Hierarchical_Purification}
\subsubsection{Mid-Risk Point Filtering}
For points in the mid-risk set $\mathcal{P}_{\text{mid}}'$, we employ gradient-guided non-uniform filtering. First, we identify the single most ``malicious'' wavelet band, denoted as $s^*$, by finding the band with the maximum gradient energy:
\begin{equation}
	\label{eq:highest_energy_band}
	s^*_i = \underset{s \in \{\frac{S}{2}, \dots, S\}}{\arg\max} \|\nabla_{\psi_{s,i}} \mathcal{L}\|_2,
\end{equation}
then, for points in $\mathcal{P}_{\text{mid}}'$, we only attenuate their wavelet coefficients at this single high-impact band:
\begin{equation}
	\label{eq:mid_risk_filtering_single}
	\psi_{s,i}' = \begin{cases}
		\gamma \psi_{s,i} & \text{if } p_i \in \mathcal{P}_{\text{mid}} \text{ and } s = s^{*}_i, \\
		\psi_{s,i}        & \text{otherwise},
	\end{cases}
\end{equation}
where $\gamma \in [0,1)$ is the attenuation factor. Applying the IGWT with these modified coefficients $\psi_{s,i}'$ reconstructs a temporary PC $\tilde{\mathcal{P}}$ via Eq. \ref{eq:igwt_reconstruction}.

\subsubsection{High-Risk Point Removal}
Points in the high-risk set $\mathcal{P}_{\text{high}}'$ are deemed hardly-recoverable outliers and are removed directly from the reconstructed temporary PC $\tilde{\mathcal{P}}$:
\begin{equation}
	\label{eq:high_risk_removal}
	\mathcal{P}'' = \tilde{\mathcal{P}} \setminus \mathcal{P}_{\text{high}},
\end{equation}
this yields the final purified PC $\mathcal{P}''$ with a reduced point count of $N - |\mathcal{P}_{\text{high}}|$.

We name our defense as Point Cloud Wavelet Purification (\defense for short), which can be used as a plug-and-play pre-processing module that can be attached to any 3D point cloud recognition model to enhance robustness.

\begin{table*}[h]
	\caption{Classification accuracy ($\uparrow$,\%) of various point cloud models with different defense methods against adversarial attacks. For each model and attack, the highest accuracy is indicated in \textbf{bold} and the second-highest in \underline{underline}.}
	\centering
	\setlength{\tabcolsep}{4pt}
	\begin{tabular}{llcccccccccc}
		\toprule
		\multirow{2}{*}{Model}      & \multirow{2}{*}{Defense} & \multicolumn{5}{c}{ModelNet40}    & \multicolumn{5}{c}{ShapeNet}                                                                                                                                                                                                                                                                                       \\
		\cmidrule(lr){3-7} \cmidrule(lr){8-12}
		                            &                          & Eidos                             & GeoA3                          & GSDA                              & HiT-ADV                        & SI-ADV                         & Eidos                          & GeoA3                             & GSDA                              & HiT-ADV                           & SI-ADV                         \\
		\midrule

		\multirow{7}{*}{CurveNet}   & No Defense               & 0.00                              & 0.00                           & 0.00                              & 7.90                           & 0.00                           & 2.02                           & 0.18                              & 3.88                              & 0.00                              & 1.10                           \\
		                            & SOR                      & 76.57                             & 69.02                          & 85.41                             & 71.05                          & 67.70                          & 87.38                          & 94.87                             & 78.09                             & 65.12                             & 87.73                          \\
		                            & ROR                      & 69.66                             & 58.78                          & 74.06                             & 68.51                          & 68.11                          & 85.31                          & 92.94                             & 75.96                             & 61.77                             & 77.33                          \\
		                            & PointCVAR                & 84.56                             & 73.33                          & 86.53                             & \underline{81.99}              & 73.48                          & \underline{96.84}              & \underline{96.86}                 & 97.58                             & \textbf{76.20}                    & \underline{97.19}              \\
		                            & PFourierP                & 86.38                             & 76.28                          & \textbf{88.35}                    & 77.17                          & \underline{78.66}              & 95.38                          & 96.81                             & \textbf{97.97}                    & 70.71                             & 97.04                          \\
		                            & Ada3Diff                 & \textbf{88.60}                    & \underline{82.35}              & 87.13                             & 66.36                          & 73.53                          & 83.82                          & 83.09                             & 59.74                             & \underline{75.37}                 & 73.71                          \\
		                            & \defense                 & \cellcolor{Gray}\underline{87.95} & \cellcolor{Gray}\textbf{82.63} & \cellcolor{Gray}\underline{87.62} & \cellcolor{Gray}\textbf{82.51} & \cellcolor{Gray}\textbf{85.42} & \cellcolor{Gray}\textbf{97.35} & \cellcolor{Gray}\textbf{97.03}    & \cellcolor{Gray}\underline{97.92} & \cellcolor{Gray}72.89             & \cellcolor{Gray}\textbf{97.74} \\
		\midrule

		\multirow{7}{*}{DGCNN}      & No Defense               & 0.00                              & 0.00                           & 0.00                              & 1.29                           & 0.18                           & 0.55                           & 0.00                              & 0.00                              & 3.68                              & 0.55                           \\
		                            & SOR                      & 66.43                             & 58.73                          & 75.54                             & 64.63                          & 61.79                          & 83.73                          & 78.23                             & 90.29                             & 70.96                             & 82.37                          \\
		                            & ROR                      & 64.65                             & 57.16                          & 70.86                             & 58.57                          & 58.46                          & 76.85                          & 78.98                             & 88.97                             & 65.41                             & 71.57                          \\
		                            & PointCVAR                & 79.69                             & 68.55                          & \underline{86.83}                 & \underline{76.24}              & 66.82                          & 91.28                          & 87.84                             & 96.23                             & 75.87                             & 89.44                          \\
		                            & PFourierP                & 82.28                             & \underline{77.20}              & 83.98                             & 63.17                          & \underline{75.68}              & 95.39                          & 90.91                             & \underline{97.10}                 & 76.23                             & \underline{92.95}              \\
		                            & Ada3Diff                 & \textbf{86.76}                    & \textbf{78.86}                 & 83.75                             & 65.07                          & \textbf{75.74}                 & \underline{95.77}              & \textbf{95.22}                    & 96.56                             & \underline{77.39}                 & 55.63                          \\
		                            & \defense                 & \cellcolor{Gray}\underline{82.30} & \cellcolor{Gray}74.45          & \cellcolor{Gray}\textbf{87.10}    & \cellcolor{Gray}\textbf{77.02} & \cellcolor{Gray}74.45          & \cellcolor{Gray}\textbf{96.12} & \cellcolor{Gray}\underline{93.72} & \cellcolor{Gray}\textbf{97.13}    & \cellcolor{Gray}\textbf{78.26}    & \cellcolor{Gray}\textbf{95.53} \\
		\midrule

		\multirow{7}{*}{PointNet}   & No Defense               & 0.00                              & 0.00                           & 0.00                              & 5.33                           & 0.00                           & 1.10                           & 0.00                              & 0.00                              & 9.41                              & 0.92                           \\
		                            & SOR                      & 75.45                             & 62.25                          & 69.35                             & 59.61                          & 63.11                          & 85.13                          & 83.01                             & 90.21                             & \underline{78.22}                 & 85.00                          \\
		                            & ROR                      & 73.13                             & 54.70                          & 61.47                             & 57.88                          & 57.58                          & 78.58                          & 79.21                             & 88.08                             & 73.10                             & 70.80                          \\
		                            & PointCVAR                & \underline{82.69}                 & \underline{75.66}              & 73.10                             & \underline{65.60}              & 65.90                          & 96.13                          & 85.81                             & 94.94                             & 74.89                             & 93.00                          \\
		                            & PFourierP                & 82.44                             & 70.55                          & \textbf{78.06}                    & 60.07                          & \underline{68.35}              & \underline{96.30}              & \underline{88.19}                 & \underline{96.86}                 & 59.92                             & \underline{94.11}              \\
		                            & Ada3Diff                 & 82.17                             & 64.34                          & 58.59                             & 63.24                          & 51.10                          & 79.96                          & 61.21                             & 64.34                             & 63.05                             & 54.41                          \\
		                            & \defense                 & \cellcolor{Gray}\textbf{83.77}    & \cellcolor{Gray}\textbf{80.24} & \cellcolor{Gray}\underline{76.85} & \cellcolor{Gray}\textbf{66.49} & \cellcolor{Gray}\textbf{75.50} & \cellcolor{Gray}\textbf{96.85} & \cellcolor{Gray}\textbf{90.25}    & \cellcolor{Gray}\textbf{97.15}    & \cellcolor{Gray}\textbf{79.67}    & \cellcolor{Gray}\textbf{94.43} \\
		\midrule

		\multirow{7}{*}{Pointnet++} & No Defense               & 0.00                              & 1.29                           & 3.44                              & 0.00                           & 0.94                           & 0.00                           & 0.00                              & 5.94                              & 0.00                              & 0.92                           \\
		                            & SOR                      & 81.44                             & 61.60                          & 81.91                             & 70.87                          & 71.09                          & 87.43                          & 77.37                             & 87.42                             & 76.92                             & 86.41                          \\
		                            & ROR                      & 73.22                             & 62.93                          & 77.47                             & 69.46                          & 63.11                          & 72.82                          & 77.07                             & 83.70                             & 71.48                             & 75.13                          \\
		                            & PointCVAR                & 88.57                             & 78.24                          & 83.72                             & \underline{81.01}              & 78.04                          & \underline{97.74}              & \textbf{97.75}                    & \underline{97.77}                 & \textbf{86.70}                    & \underline{97.22}              \\
		                            & PFourierP                & 85.03                             & 80.15                          & \underline{84.04}                 & 71.97                          & \underline{78.43}              & 96.03                          & 91.46                             & 97.73                             & 71.50                             & 95.59                          \\
		                            & Ada3Diff                 & \underline{88.97}                 & \underline{80.88}              & 80.31                             & 75.74                          & 55.62                          & 97.24                          & 95.22                             & 93.44                             & 72.79                             & 69.85                          \\
		                            & \defense                 & \cellcolor{Gray}\textbf{89.81}    & \cellcolor{Gray}\textbf{85.10} & \cellcolor{Gray}\textbf{89.05}    & \cellcolor{Gray}\textbf{82.07} & \cellcolor{Gray}\textbf{83.68} & \cellcolor{Gray}\textbf{98.08} & \cellcolor{Gray}\underline{96.10} & \cellcolor{Gray}\textbf{98.12}    & \cellcolor{Gray}\underline{86.46} & \cellcolor{Gray}\textbf{98.68} \\
		\bottomrule
	\end{tabular}
	\label{tab:exp-main-compact}
\end{table*}

\section{Experiments}
\label{Experiments}
\subsection{Setup}
\subsubsection{Dataset and Model}
Our experiments are conducted based on ModelNet40 \cite{wu20153d} and ShapeNet \cite{chang2015shapenet} for classification task, and we utilize DGCNN \cite{10.1145/3326362}, PointNet \cite{8099499}, Pointnet++ \cite{qi2017pointnetdeephierarchicalfeature} and CurveNet \cite{xiang2021walk} as the victim models.

\subsubsection{Attack Methods}
We consider five attack methods: GSDA \cite{liu2023pointcloudattacksgraph}, GeoA3 \cite{9294112}, HiT-ADV \cite{lou2024hide}, SI-ADV \cite{huang2022shape}, and Eidos \cite{sicre2024eidos}.

\subsubsection{Baselines}
At present, few methods are compatible with our defense setting. We compare our method against several baselines, including two classic techniques: \textbf{Statistical Outlier Removal (SOR)} \cite{sor} and \textbf{Radius Outlier Removal (ROR)} \cite{ror}. Additionally, we evaluate against more advanced methods:
\begin{itemize}[leftmargin=*]
\item \textbf{PointCVAR} \cite{li2024pointcvar} is a saliency-based perturbation removal method that leverages tail risk minimization principles to effectively filter diverse outliers.
\end{itemize}
We also designed a custom baseline:
\begin{itemize}[leftmargin=*]
\item \textbf{PFourierP}, which performs Graph Fourier Transform (GFT)-based low-pass filtering through hard truncation. Based on our search, the optimal cutoff was found to be $\approx 0.67$, defining the filter as $\varphi(\lambda_k)=1$ if $\lambda_k \leq 0.67$, and $0$ otherwise.
\end{itemize}
To better position \defense among state-of-the-art defenses, we include a training-based purification defense, through it doesn't match our threat model:
\begin{itemize}[leftmargin=*]
\item \textbf{Ada3Diff} \cite{zhang2023ada3diff} is a diffusion model–based defense mechanism that purifies adversarial 3D point clouds by first estimating the distortion level of the input and then adaptively selecting an appropriate diffusion timestep to guide the reverse denoising process, thereby restoring the underlying clean data distribution. As the official code is not available, we re-implemented this method ourselves.
\end{itemize}

\begin{figure}[h]
	\centering
	\includegraphics[width=1\linewidth]{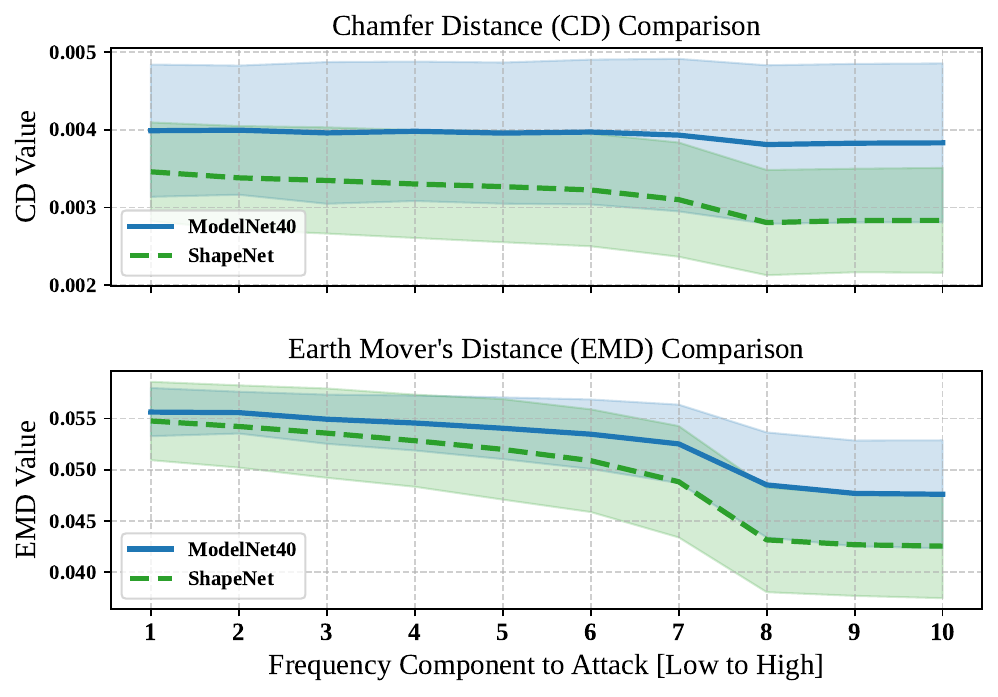}
	\caption{Impact of attack frequency on distance metrics. Under a fixed perturbation energy ($\|\Delta\|_F=2$), the CD are basically stable, while EMD decreases significantly as the attack frequency increases, match our theoretical analysis.}
	\label{fig:quantitative_results}
\end{figure}

\begin{figure}[h]
	\centering
	\includegraphics[width=0.7\linewidth]{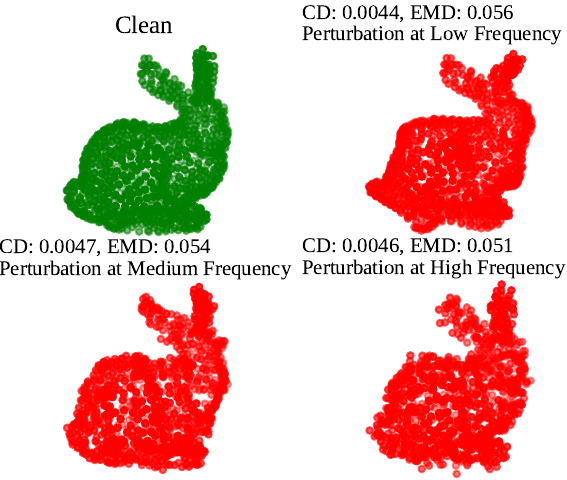}
	\caption{Qualitative comparison of perturbations. The high-frequency attack results in less visual distortion and a lower EMD than the low-frequency attack.}
	\label{fig:visualization_results}
\end{figure}
\subsubsection{Implementation Details}
We set $\alpha = 0.002$, $\beta = 1$, and $\gamma = 0$, and use the classification accuracy of the purified point cloud as the evaluation metric, following \cite{li2024pointcvar}. All reported results are averaged over five independent runs.

\subsection{Main Results}
\label{subsec:main_results}
\subsubsection{Spectral Analysis}
In order to experimentally verify our theoretical analysis of the characteristics of imperceptible perturbations, we conduct a numerical validation. We first decompose each PC into its spectral domain and divide the spectrum into 10 uniform frequency bands, from low to high  ($\lambda_0$ to $\lambda_{max}$). We then inject a perturbation with the same fixed energy ($\|\Delta\|_F=2$) into each band and measure the resulting CD and EMD against the original PC. The experimental results, shown in Figure~\ref{fig:quantitative_results}, clearly support our  theoretical analysis. The CD value remains nearly constant across all frequency bands, indicates its connection to $\|\Delta\|_F$. Crucially, the EMD value exhibits a clear downward trend as the perturbated frequency increases. This indicates that, for the same amount of energy, high-frequency perturbations indeed result in a smaller perceptual distance (i.e., are less noticeable) than low-frequency ones.
Figure~\ref{fig:visualization_results} further provides an intuitive confirmation. The low-frequency perturbation causes a visible distortion of the shape, whereas the high-frequency one resembles fine-grained surface noise, better preserving the geometry.

\begin{table*}[ht]
	\centering
	\caption{Evaluation of different purification methods against SI-ADV attacks. The values represent the Chamfer Distance (CD $\downarrow$) between the original PCs and attacked/purified PCs. A lower CD value indicates better preservation of geometric information.}
	\begin{tabular}{ccccccccc}
		\toprule
		Model    & Dataset    & \underline{\textit{SI-ADV}} & SOR      & ROR      & PFourierP & PointCVAR         & Ada3Diff          & \defense          \\
		\midrule
		DGCNN    & ModelNet40 & 1.52e-03                    & 2.67e-03 & 5.81e-03 & 2.71e-03  & 1.47e-03          & \textbf{6.12e-04} & 1.43e-03          \\
		PointNet & ModelNet40 & 5.84e-04                    & 1.60e-03 & 4.79e-03 & 2.16e-03  & \textbf{4.68e-04} & 5.25e-04          & 5.37e-04          \\
		DGCNN    & ShapeNet   & 1.76e-03                    & 2.39e-03 & 3.42e-03 & 2.64e-03  & 1.63e-03          & 2.35e-03          & \textbf{1.60e-03} \\
		PointNet & ShapeNet   & 8.14e-04                    & 1.06e-03 & 2.44e-03 & 2.07e-03  & 5.87e-04          & 7.06e-04          & \textbf{3.68e-04} \\
		\bottomrule
	\end{tabular}
	\label{tab:cd-eval}
\end{table*}

\subsubsection{Quantitative Evaluation}
As shown in Table \ref{tab:exp-main-compact}, \defense demonstrates consistently superior performance across all evaluated models, datasets, and adversarial attacks. Notably, \defense significantly outperforms both classic baselines like SOR and ROR, as well as more advanced methods like PointCVAR and our custom GFT-based filter, PFourierP. In contrast, Ada3Diff, another state-of-the-art diffusion-based defense, exhibits unstable performance across different attacks. This instability likely stems from its reliance on a distortion estimation module, which may fail to accurately gauge the severity of sophisticated, imperceptible attacks, leading to the selection of a suboptimal purification intensity. These results validate the effectiveness of our hierarchical purification strategy, which leverages graph wavelets to precisely identify and mitigate adversarial perturbations while preserving the geometric integrity of the PCs. Besides, the remarkable performance of PFourierP and \defense seems to suggest that relying solely on deleting suspicious points may not yield the desired results. Furthermore, PFourierP, as a simple low-pass filter, also demonstrated impressive performance, thus supporting our insight that the imperceptible perturbations are mainly located at high frequencies.

To further evaluate our method's ability to preserve geometric information, we report the CD between the clean-attacked/purified PCs. As shown in Table \ref{tab:cd-eval}, our method is benchmarked against several defenses on point clouds adversarially attacked by SI-ADV. The results demonstrate that our approach achieves significantly lower CD values. This indicates that our method not only removes adversarial perturbations but also better preserves the underlying geometry of the original PC compared to other methods.

\subsubsection{Discussion}
\begin{figure}[h]
	\centering
	\includegraphics[width=1\linewidth]{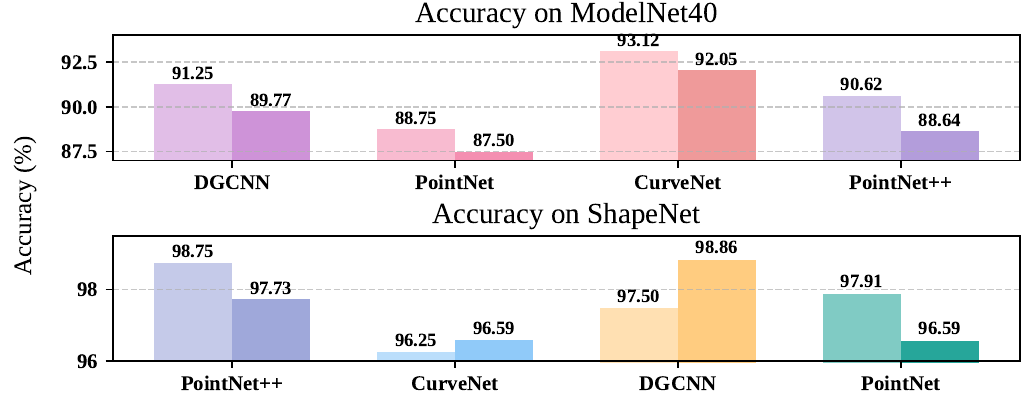}
	\caption{Side-effect Study of \defense on Clean Data: The left bar in each group is the accuracy ($\uparrow$,\%) on clean samples; the right bar is the accuracy after ``purification".}
	\label{fig:side_effect}
\end{figure}

A crucial requirement for any practical defense mechanism is that it must not unduly penalize the model's performance on benign, non-adversarial data. A defense that significantly degrades accuracy on clean inputs would be impractical for real-world deployment. Therefore, we conducted a study to evaluate the side-effects of \defense by applying it to original, unmodified samples. As shown in Figure~\ref{fig:side_effect}, our method causes only a minimal performance drop of approximately 1\%. This negligible impact is critical, as it demonstrates that \defense strikes an excellent trade-off between adversarial robustness and benign accuracy, ensuring its viability for practical applications.

In addition to modifying the coordinates of existing points, an attacker can also add extra malicious points to the clean PC. Thus, following the settings of \cite{li2024pointcvar}, we implement the point addition attacks, which add points by Chamfer Distance (CD).
As shown in Table \ref{tab:pointadd}, this attack severely degrades the performance of undefended models. Our proposed method, \defense, demonstrates remarkable robustness.

\begin{figure*}[h]
	\centering
	\includegraphics[width=1\linewidth]{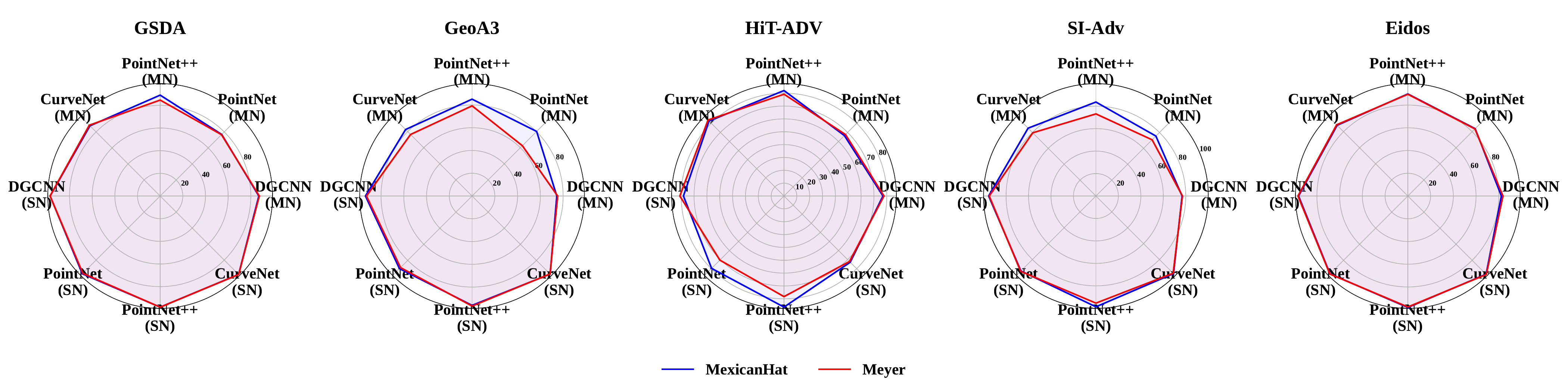}
	\caption{Comparing the performance of \defense using the Meyer wavelet versus the Mexican Hat wavelet on ModelNet40 (MN) and ShapeNet (SN) datasets. The Mexican Hat consistently achieves superior or comparable performance.}
	\label{fig:radar_charts_kernel}
\end{figure*}
\begin{figure}[h]
	\centering
	\includegraphics[width=1\linewidth]{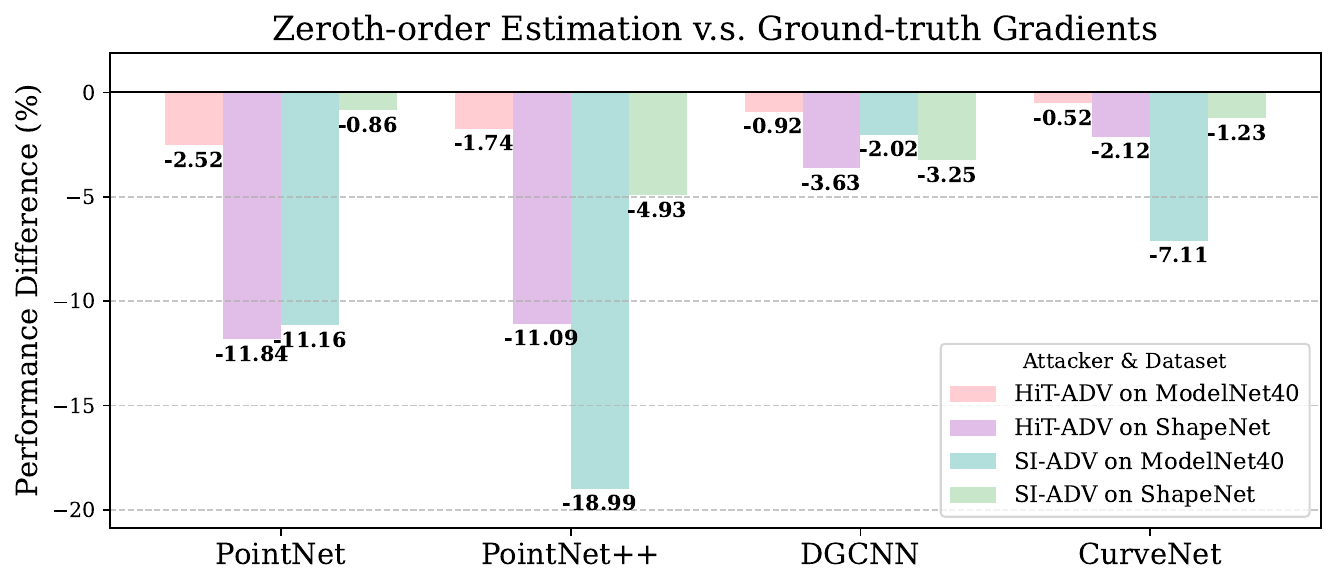}
	\caption{Impact on model accuracy for black-box defense. The bars show the absolute difference in accuracy between a defense guided by approximated gradients and one guided by ground-truth gradients.}
	\label{fig:zeroth_gard}
\end{figure}
In certain real-world scenarios, our defender may not have access to the model's gradients or feature maps, resulting in a black-box defense setting. Regarding this, we can approximate the loss gradient $\mathcal{L}_{CE}$ using a zeroth-order estimation while ignoring the intermediate feature term $\mathcal{F}_l$. As shown in Figure \ref{fig:zeroth_gard}, this approach proves generally effective, showing an acceptable performance degradation in most cases (avg. -$5.24\%$) compared to the ground-truth gradient. However, its performance appears to be model-dependent, with a significant accuracy drop observed for the PointNet++ (up to -$18.99\%$). This suggests that while the zeroth-order approximation is a viable strategy for black-box defense, caution should be exercised, as its effectiveness may vary across different model architectures.

\begin{table}[h]
	\centering
	\caption{Classification accuracy ($\uparrow$,\%) comparison under the point addition attack on ModelNet40 and ShapeNet. Our method (PWaveP) demonstrates superior defensive performance over the baseline PointCVAR. The best results are highlighted in \textbf{bold}.}
	\begin{tabular}{ccccc}
		\toprule
		\textbf{Model} & \multicolumn{1}{c}{\textbf{Dataset}} & \textbf{Clean} & \multicolumn{1}{c}{\textbf{PointCVAR}} & \textbf{PWaveP} \\ \midrule
		DGCNN          & Modelnet40                           & 0.28           & 84.24                                  & \textbf{87.25}  \\
		DGCNN          & ShapeNet                             & 0.52           & 95.48                                  & \textbf{96.24}  \\
		PointNet       & Modelnet40                           & 2.31           & 88.57                                  & \textbf{91.06}  \\
		PointNet       & ShapeNet                             & 3.62           & 97.70                                  & \textbf{98.47}  \\ \bottomrule
	\end{tabular}
	\label{tab:pointadd}
\end{table}

\subsection{Ablation Study}
\label{subsec:ablation_study}
\subsubsection{Impact of Kernel Selection}
To analyze the effect of the wavelet kernel, we evaluated our defense performance using the Mexican Hat or Meyer wavelets. The results, presented in Figure~\ref{fig:radar_charts_kernel}, demonstrate that the Mexican Hat kernel performs better. While the Meyer wavelet offers theoretical elegance, the structure of the Mexican Hat kernel is better suited for isolating the high-frequency components characteristic of adversarial perturbations, thus leading to superior performance in practice.

\begin{figure}[h]
	\includegraphics[width=0.95\linewidth]{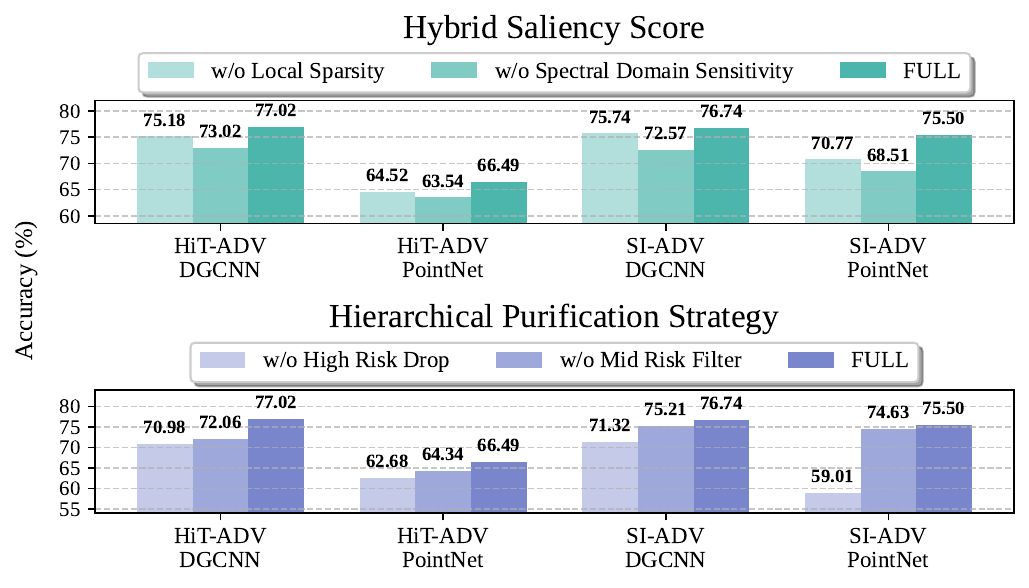}
	\caption{Ablation study validating the contributions on ModelNet40 of each component in our hybrid saliency score (top) and our hierarchical purification strategy (bottom).}
	\label{fig:ablation}
\end{figure}
\subsubsection{Contribution of Core Modules}
We conducted an ablation study to validate the contribution of our method's key components. We individually evaluated the parts of our hybrid saliency score (spectral and spatial) and our hierarchical purification strategy (filtering and removal). As shown in Figure~\ref{fig:ablation}, removing any component led to a noticeable degradation in performance, confirming that each part of our proposed component is contributes positively to the final result.



\section{Conclusion}
In this paper, we addressed the threat of imperceptible adversarial perturbations on 3D point clouds by first establishing, both theoretically and empirically, that such perturbations primarily manifest as high-frequency noise in the graph spectral domain. Building on this core finding, we introduced \defense, a non-invasive, training-free and plug-and-play purification framework. Our method leverages the excellent spatial-spectral analysis and employs a hierarchical strategy, guided by a hybrid saliency score, to effectively filter moderately perturbed points and remove severe outliers. Comprehensive experiments validate that \defense offering a practical solution to enhance the security of 3D web systems.

\section*{Acknowledgments}
This research was funded in part by the National Natural Science Foundation of China under Grant 62372096.

\clearpage
\bibliographystyle{ACM-Reference-Format}
\balance 
\bibliography{sample-base}


\begin{thebibliography}{48}


\ifx \showCODEN    \undefined \def \showCODEN     #1{\unskip}     \fi
\ifx \showISBNx    \undefined \def \showISBNx     #1{\unskip}     \fi
\ifx \showISBNxiii \undefined \def \showISBNxiii  #1{\unskip}     \fi
\ifx \showISSN     \undefined \def \showISSN      #1{\unskip}     \fi
\ifx \showLCCN     \undefined \def \showLCCN      #1{\unskip}     \fi
\ifx \shownote     \undefined \def \shownote      #1{#1}          \fi
\ifx \showarticletitle \undefined \def \showarticletitle #1{#1}   \fi
\ifx \showURL      \undefined \def \showURL       {\relax}        \fi
\providecommand\bibfield[2]{#2}
\providecommand\bibinfo[2]{#2}
\providecommand\natexlab[1]{#1}
\providecommand\showeprint[2][]{arXiv:#2}

\bibitem[Benson et~al\mbox{.}(2016)]%
        {benson2016higher}
\bibfield{author}{\bibinfo{person}{Austin~R Benson}, \bibinfo{person}{David~F
  Gleich}, {and} \bibinfo{person}{Jure Leskovec}.}
  \bibinfo{year}{2016}\natexlab{}.
\newblock \showarticletitle{Higher-order organization of complex networks}.
\newblock \bibinfo{journal}{\emph{Science}} (\bibinfo{year}{2016}).
\newblock


\bibitem[Chang et~al\mbox{.}(2015)]%
        {chang2015shapenet}
\bibfield{author}{\bibinfo{person}{Angel~X Chang}, \bibinfo{person}{Thomas
  Funkhouser}, \bibinfo{person}{Leonidas Guibas}, \bibinfo{person}{Pat
  Hanrahan}, \bibinfo{person}{Qixing Huang}, \bibinfo{person}{Zimo Li},
  \bibinfo{person}{Silvio Savarese}, \bibinfo{person}{Manolis Savva},
  \bibinfo{person}{Shuran Song}, \bibinfo{person}{Hao Su}, {et~al\mbox{.}}}
  \bibinfo{year}{2015}\natexlab{}.
\newblock \showarticletitle{Shapenet: An information-rich 3d model repository}.
\newblock \bibinfo{journal}{\emph{arXiv}} (\bibinfo{year}{2015}).
\newblock


\bibitem[Charles et~al\mbox{.}(2017)]%
        {8099499}
\bibfield{author}{\bibinfo{person}{R.~Qi Charles}, \bibinfo{person}{Hao Su},
  \bibinfo{person}{Mo Kaichun}, {and} \bibinfo{person}{Leonidas~J. Guibas}.}
  \bibinfo{year}{2017}\natexlab{}.
\newblock \showarticletitle{PointNet: Deep Learning on Point Sets for 3D
  Classification and Segmentation}. In \bibinfo{booktitle}{\emph{CVPR}}.
\newblock


\bibitem[Chen et~al\mbox{.}(2017)]%
        {chen2017multi}
\bibfield{author}{\bibinfo{person}{Xiaozhi Chen}, \bibinfo{person}{Huimin Ma},
  \bibinfo{person}{Ji Wan}, \bibinfo{person}{Bo Li}, {and}
  \bibinfo{person}{Tian Xia}.} \bibinfo{year}{2017}\natexlab{}.
\newblock \showarticletitle{Multi-view 3d object detection network for
  autonomous driving}. In \bibinfo{booktitle}{\emph{CVPR}}.
\newblock


\bibitem[Chung(1997)]%
        {chung1997spectral}
\bibfield{author}{\bibinfo{person}{Fan~RK Chung}.}
  \bibinfo{year}{1997}\natexlab{}.
\newblock \bibinfo{booktitle}{\emph{Spectral graph theory}}.
  Vol.~\bibinfo{volume}{92}.
\newblock \bibinfo{publisher}{American Mathematical Soc.}
\newblock


\bibitem[Defferrard et~al\mbox{.}(2016)]%
        {defferrard2016convolutional}
\bibfield{author}{\bibinfo{person}{Micha{\"e}l Defferrard},
  \bibinfo{person}{Xavier Bresson}, {and} \bibinfo{person}{Pierre
  Vandergheynst}.} \bibinfo{year}{2016}\natexlab{}.
\newblock \showarticletitle{Convolutional neural networks on graphs with fast
  localized spectral filtering}. In \bibinfo{booktitle}{\emph{NeurIPS}}.
\newblock


\bibitem[Dong et~al\mbox{.}(2020)]%
        {dong2020self}
\bibfield{author}{\bibinfo{person}{Xiaoyi Dong}, \bibinfo{person}{Dongdong
  Chen}, \bibinfo{person}{Hang Zhou}, \bibinfo{person}{Gang Hua},
  \bibinfo{person}{Weiming Zhang}, {and} \bibinfo{person}{Nenghai Yu}.}
  \bibinfo{year}{2020}\natexlab{}.
\newblock \showarticletitle{Self-robust 3d point recognition via gather-vector
  guidance}. In \bibinfo{booktitle}{\emph{CVPR}}.
\newblock


\bibitem[Fan et~al\mbox{.}(2024)]%
        {fan2024invisible}
\bibfield{author}{\bibinfo{person}{Linkun Fan}, \bibinfo{person}{Fazhi He},
  \bibinfo{person}{Tongzhen Si}, \bibinfo{person}{Wei Tang}, {and}
  \bibinfo{person}{Bing Li}.} \bibinfo{year}{2024}\natexlab{}.
\newblock \showarticletitle{Invisible backdoor attack against 3D point cloud
  classifier in graph spectral domain}. In \bibinfo{booktitle}{\emph{AAAI}}.
\newblock


\bibitem[Feng et~al\mbox{.}(2023)]%
        {10.1093/comjnl/bxad109}
\bibfield{author}{\bibinfo{person}{Le Feng}, \bibinfo{person}{Zhenxing Qian},
  \bibinfo{person}{Xinpeng Zhang}, {and} \bibinfo{person}{Sheng Li}.}
  \bibinfo{year}{2023}\natexlab{}.
\newblock \showarticletitle{Stealthy Backdoor Attacks On Deep Point Cloud
  Recognization Networks}.
\newblock \bibinfo{journal}{\emph{Comput. J.}}  \bibinfo{volume}{67}
  (\bibinfo{year}{2023}).
\newblock


\bibitem[Gao et~al\mbox{.}(2023)]%
        {gao2023imperceptible}
\bibfield{author}{\bibinfo{person}{Kuofeng Gao}, \bibinfo{person}{Jiawang Bai},
  \bibinfo{person}{Baoyuan Wu}, \bibinfo{person}{Mengxi Ya}, {and}
  \bibinfo{person}{Shu-Tao Xia}.} \bibinfo{year}{2023}\natexlab{}.
\newblock \showarticletitle{Imperceptible and robust backdoor attack in 3d
  point cloud}.
\newblock \bibinfo{journal}{\emph{IEEE Transactions on Information Forensics
  and Security}}  \bibinfo{volume}{19} (\bibinfo{year}{2023}).
\newblock


\bibitem[Guo et~al\mbox{.}(2020)]%
        {guo2020deep}
\bibfield{author}{\bibinfo{person}{Yulan Guo}, \bibinfo{person}{Hanyun Wang},
  \bibinfo{person}{Qingyong Hu}, \bibinfo{person}{Hao Liu}, \bibinfo{person}{Li
  Liu}, {and} \bibinfo{person}{Mohammed Bennamoun}.}
  \bibinfo{year}{2020}\natexlab{}.
\newblock \showarticletitle{Deep learning for 3d point clouds: A survey}.
\newblock \bibinfo{journal}{\emph{IEEE transactions on pattern analysis and
  machine intelligence}}  \bibinfo{volume}{43} (\bibinfo{year}{2020}).
\newblock


\bibitem[Hammond et~al\mbox{.}(2011)]%
        {HAMMOND2011129}
\bibfield{author}{\bibinfo{person}{David~K. Hammond}, \bibinfo{person}{Pierre
  Vandergheynst}, {and} \bibinfo{person}{Rémi Gribonval}.}
  \bibinfo{year}{2011}\natexlab{}.
\newblock \showarticletitle{Wavelets on graphs via spectral graph theory}.
\newblock \bibinfo{journal}{\emph{Applied and Computational Harmonic Analysis}}
   \bibinfo{volume}{30} (\bibinfo{year}{2011}).
\newblock


\bibitem[Huang et~al\mbox{.}(2022)]%
        {huang2022shape}
\bibfield{author}{\bibinfo{person}{Qidong Huang}, \bibinfo{person}{Xiaoyi
  Dong}, \bibinfo{person}{Dongdong Chen}, \bibinfo{person}{Hang Zhou},
  \bibinfo{person}{Weiming Zhang}, {and} \bibinfo{person}{Nenghai Yu}.}
  \bibinfo{year}{2022}\natexlab{}.
\newblock \showarticletitle{Shape-invariant 3D adversarial point clouds}. In
  \bibinfo{booktitle}{\emph{CVPR}}.
\newblock


\bibitem[Huang et~al\mbox{.}(2024)]%
        {huang2024pointcat}
\bibfield{author}{\bibinfo{person}{Qidong Huang}, \bibinfo{person}{Xiaoyi
  Dong}, \bibinfo{person}{Dongdong Chen}, \bibinfo{person}{Hang Zhou},
  \bibinfo{person}{Weiming Zhang}, \bibinfo{person}{Kui Zhang},
  \bibinfo{person}{Gang Hua}, \bibinfo{person}{Yueqiang Cheng}, {and}
  \bibinfo{person}{Nenghai Yu}.} \bibinfo{year}{2024}\natexlab{}.
\newblock \showarticletitle{PointCAT: Contrastive adversarial training for
  robust point cloud recognition}.
\newblock \bibinfo{journal}{\emph{IEEE Transactions on Image Processing}}
  (\bibinfo{year}{2024}).
\newblock


\bibitem[Li et~al\mbox{.}(2022)]%
        {li2022robust}
\bibfield{author}{\bibinfo{person}{Kaidong Li}, \bibinfo{person}{Ziming Zhang},
  \bibinfo{person}{Cuncong Zhong}, {and} \bibinfo{person}{Guanghui Wang}.}
  \bibinfo{year}{2022}\natexlab{}.
\newblock \showarticletitle{Robust structured declarative classifiers for 3d
  point clouds: Defending adversarial attacks with implicit gradients}. In
  \bibinfo{booktitle}{\emph{CVPR}}.
\newblock


\bibitem[Li et~al\mbox{.}(2024)]%
        {li2024pointcvar}
\bibfield{author}{\bibinfo{person}{Xinke Li}, \bibinfo{person}{Junchi Lu},
  \bibinfo{person}{Henghui Ding}, \bibinfo{person}{Changsheng Sun},
  \bibinfo{person}{Joey~Tianyi Zhou}, {and} \bibinfo{person}{Yeow~Meng Chee}.}
  \bibinfo{year}{2024}\natexlab{}.
\newblock \showarticletitle{PointCVaR: Risk-optimized outlier removal for
  robust 3D point cloud classification}. In \bibinfo{booktitle}{\emph{AAAI}}.
\newblock


\bibitem[Liu and Hu(2024)]%
        {liu2024explicitly}
\bibfield{author}{\bibinfo{person}{Daizong Liu} {and} \bibinfo{person}{Wei
  Hu}.} \bibinfo{year}{2024}\natexlab{}.
\newblock \showarticletitle{Explicitly perceiving and preserving the local
  geometric structures for 3d point cloud attack}. In
  \bibinfo{booktitle}{\emph{AAAI}}.
\newblock


\bibitem[Liu et~al\mbox{.}(2023a)]%
        {liu2023pointcloudattacksgraph}
\bibfield{author}{\bibinfo{person}{Daizong Liu}, \bibinfo{person}{Wei Hu},
  {and} \bibinfo{person}{Xin Li}.} \bibinfo{year}{2023}\natexlab{a}.
\newblock \showarticletitle{Point Cloud Attacks in Graph Spectral Domain: When
  3D Geometry Meets Graph Signal Processing}.
\newblock \bibinfo{journal}{\emph{arXiv}} (\bibinfo{year}{2023}).
\newblock


\bibitem[Liu et~al\mbox{.}(2019)]%
        {liu2019extending}
\bibfield{author}{\bibinfo{person}{Daniel Liu}, \bibinfo{person}{Ronald Yu},
  {and} \bibinfo{person}{Hao Su}.} \bibinfo{year}{2019}\natexlab{}.
\newblock \showarticletitle{Extending adversarial attacks and defenses to deep
  3d point cloud classifiers}. In \bibinfo{booktitle}{\emph{IEEE International
  Conference on Image Processing (ICIP)}}.
\newblock


\bibitem[Liu et~al\mbox{.}(2025)]%
        {liu2025stba}
\bibfield{author}{\bibinfo{person}{Renyang Liu}, \bibinfo{person}{Kwok-Yan
  Lam}, \bibinfo{person}{Wei Zhou}, \bibinfo{person}{Sixing Wu},
  \bibinfo{person}{Jun Zhao}, \bibinfo{person}{Dongting Hu}, {and}
  \bibinfo{person}{Mingming Gong}.} \bibinfo{year}{2025}\natexlab{}.
\newblock \showarticletitle{STBA: Towards Evaluating the Robustness of DNNs for
  Query-Limited Black-box Scenario}.
\newblock \bibinfo{journal}{\emph{IEEE Transactions on Multimedia}}
  (\bibinfo{year}{2025}).
\newblock


\bibitem[Liu et~al\mbox{.}(2023b)]%
        {liu2023model}
\bibfield{author}{\bibinfo{person}{Renyang Liu}, \bibinfo{person}{Wei Zhou},
  \bibinfo{person}{Jinhong Zhang}, \bibinfo{person}{Xiaoyuan Liu},
  \bibinfo{person}{Peiyuan Si}, {and} \bibinfo{person}{Haoran Li}.}
  \bibinfo{year}{2023}\natexlab{b}.
\newblock \showarticletitle{Model Inversion Attacks on Homogeneous and
  Heterogeneous Graph Neural Networks}. In
  \bibinfo{booktitle}{\emph{SecureComm}}.
\newblock


\bibitem[Lou et~al\mbox{.}(2024)]%
        {lou2024hide}
\bibfield{author}{\bibinfo{person}{Tianrui Lou}, \bibinfo{person}{Xiaojun Jia},
  \bibinfo{person}{Jindong Gu}, \bibinfo{person}{Li Liu},
  \bibinfo{person}{Siyuan Liang}, \bibinfo{person}{Bangyan He}, {and}
  \bibinfo{person}{Xiaochun Cao}.} \bibinfo{year}{2024}\natexlab{}.
\newblock \showarticletitle{Hide in thicket: Generating imperceptible and
  rational adversarial perturbations on 3d point clouds}. In
  \bibinfo{booktitle}{\emph{CVPR}}.
\newblock


\bibitem[Ma et~al\mbox{.}(2020)]%
        {ma2020efficient}
\bibfield{author}{\bibinfo{person}{Chengcheng Ma}, \bibinfo{person}{Weiliang
  Meng}, \bibinfo{person}{Baoyuan Wu}, \bibinfo{person}{Shibiao Xu}, {and}
  \bibinfo{person}{Xiaopeng Zhang}.} \bibinfo{year}{2020}\natexlab{}.
\newblock \showarticletitle{Efficient joint gradient based attack against sor
  defense for 3d point cloud classification}. In
  \bibinfo{booktitle}{\emph{MM}}.
\newblock


\bibitem[Ma et~al\mbox{.}(2022)]%
        {ma2022rethinkingnetworkdesignlocal}
\bibfield{author}{\bibinfo{person}{Xu Ma}, \bibinfo{person}{Can Qin},
  \bibinfo{person}{Haoxuan You}, \bibinfo{person}{Haoxi Ran}, {and}
  \bibinfo{person}{Yun Fu}.} \bibinfo{year}{2022}\natexlab{}.
\newblock \showarticletitle{Rethinking Network Design and Local Geometry in
  Point Cloud: A Simple Residual MLP Framework}.
\newblock \bibinfo{journal}{\emph{arXiv}} (\bibinfo{year}{2022}).
\newblock


\bibitem[Qi et~al\mbox{.}(2017a)]%
        {qi2017pointnet}
\bibfield{author}{\bibinfo{person}{Charles~R Qi}, \bibinfo{person}{Hao Su},
  \bibinfo{person}{Kaichun Mo}, {and} \bibinfo{person}{Leonidas~J Guibas}.}
  \bibinfo{year}{2017}\natexlab{a}.
\newblock \showarticletitle{Pointnet: Deep learning on point sets for 3d
  classification and segmentation}. In \bibinfo{booktitle}{\emph{CVPR}}.
  \bibinfo{pages}{652--660}.
\newblock


\bibitem[Qi et~al\mbox{.}(2017b)]%
        {qi2017pointnetdeephierarchicalfeature}
\bibfield{author}{\bibinfo{person}{Charles~R. Qi}, \bibinfo{person}{Li Yi},
  \bibinfo{person}{Hao Su}, {and} \bibinfo{person}{Leonidas~J. Guibas}.}
  \bibinfo{year}{2017}\natexlab{b}.
\newblock \showarticletitle{PointNet++: Deep Hierarchical Feature Learning on
  Point Sets in a Metric Space}.
\newblock \bibinfo{journal}{\emph{arXiv}} (\bibinfo{year}{2017}).
\newblock


\bibitem[Rusu(2010)]%
        {sor}
\bibfield{author}{\bibinfo{person}{Radu~Bogdan Rusu}.}
  \bibinfo{year}{2010}\natexlab{}.
\newblock \showarticletitle{Semantic 3D object maps for everyday manipulation
  in human living environments}.
\newblock \bibinfo{journal}{\emph{KI-K{\"u}nstliche Intelligenz}}
  (\bibinfo{year}{2010}).
\newblock


\bibitem[Rusu and Cousins(2011)]%
        {ror}
\bibfield{author}{\bibinfo{person}{Radu~Bogdan Rusu} {and}
  \bibinfo{person}{Steve Cousins}.} \bibinfo{year}{2011}\natexlab{}.
\newblock \showarticletitle{3d is here: Point cloud library (pcl)}. In
  \bibinfo{booktitle}{\emph{ICRA}}.
\newblock


\bibitem[Rusu et~al\mbox{.}(2008)]%
        {rusu2008towards}
\bibfield{author}{\bibinfo{person}{Radu~Bogdan Rusu},
  \bibinfo{person}{Zoltan~Csaba Marton}, \bibinfo{person}{Nico Blodow},
  \bibinfo{person}{Mihai Dolha}, {and} \bibinfo{person}{Michael Beetz}.}
  \bibinfo{year}{2008}\natexlab{}.
\newblock \showarticletitle{Towards 3D point cloud based object maps for
  household environments}.
\newblock \bibinfo{journal}{\emph{Robotics and Autonomous Systems}}
  (\bibinfo{year}{2008}).
\newblock


\bibitem[Shuman et~al\mbox{.}(2013)]%
        {6494675}
\bibfield{author}{\bibinfo{person}{David~I Shuman}, \bibinfo{person}{Sunil~K.
  Narang}, \bibinfo{person}{Pascal Frossard}, \bibinfo{person}{Antonio Ortega},
  {and} \bibinfo{person}{Pierre Vandergheynst}.}
  \bibinfo{year}{2013}\natexlab{}.
\newblock \showarticletitle{The emerging field of signal processing on graphs:
  Extending high-dimensional data analysis to networks and other irregular
  domains}.
\newblock \bibinfo{journal}{\emph{IEEE Signal Processing Magazine}}
  \bibinfo{volume}{30} (\bibinfo{year}{2013}).
\newblock


\bibitem[Sicre et~al\mbox{.}(2024)]%
        {sicre2024eidos}
\bibfield{author}{\bibinfo{person}{Ronan Sicre}, \bibinfo{person}{Xiaowei
  Huang}, \bibinfo{person}{Holger Hermanns}, {and} \bibinfo{person}{Lijun
  Zhang}.} \bibinfo{year}{2024}\natexlab{}.
\newblock \showarticletitle{Eidos: Efficient, Imperceptible Adversarial 3D}. In
  \bibinfo{booktitle}{\emph{SETTA 2024}}.
\newblock


\bibitem[Sun et~al\mbox{.}(2021)]%
        {sun2021adversarially}
\bibfield{author}{\bibinfo{person}{Jiachen Sun}, \bibinfo{person}{Yulong Cao},
  \bibinfo{person}{Christopher~B Choy}, \bibinfo{person}{Zhiding Yu},
  \bibinfo{person}{Anima Anandkumar}, \bibinfo{person}{Zhuoqing~Morley Mao},
  {and} \bibinfo{person}{Chaowei Xiao}.} \bibinfo{year}{2021}\natexlab{}.
\newblock \showarticletitle{Adversarially robust 3d point cloud recognition
  using self-supervisions}. In \bibinfo{booktitle}{\emph{NeurIPS}}.
\newblock


\bibitem[Sun et~al\mbox{.}(2022)]%
        {sun2022pointdp}
\bibfield{author}{\bibinfo{person}{Jiachen Sun}, \bibinfo{person}{Weili Nie},
  \bibinfo{person}{Zhiding Yu}, \bibinfo{person}{Z~Morley Mao}, {and}
  \bibinfo{person}{Chaowei Xiao}.} \bibinfo{year}{2022}\natexlab{}.
\newblock \showarticletitle{PointDP: Diffusion-driven purification against
  adversarial attacks on 3D point cloud recognition}.
\newblock \bibinfo{journal}{\emph{arXiv preprint arXiv:2208.09801}}
  (\bibinfo{year}{2022}).
\newblock


\bibitem[Tang et~al\mbox{.}(2025)]%
        {Tang_Du_Peng_Wang_Liu_Liu_Tian_2025}
\bibfield{author}{\bibinfo{person}{Keke Tang}, \bibinfo{person}{Ziyong Du},
  \bibinfo{person}{Weilong Peng}, \bibinfo{person}{Xiaofei Wang},
  \bibinfo{person}{Daizong Liu}, \bibinfo{person}{Ligang Liu}, {and}
  \bibinfo{person}{Zhihong Tian}.} \bibinfo{year}{2025}\natexlab{}.
\newblock \showarticletitle{Imperceptible 3D Point Cloud Attacks on
  Lattice-based Barycentric Coordinates}.
\newblock \bibinfo{journal}{\emph{AAAI}} (\bibinfo{year}{2025}).
\newblock


\bibitem[Tang et~al\mbox{.}(2022)]%
        {tang2022normalattack}
\bibfield{author}{\bibinfo{person}{Keke Tang}, \bibinfo{person}{Yawen Shi},
  \bibinfo{person}{Jianpeng Wu}, \bibinfo{person}{Weilong Peng},
  \bibinfo{person}{Asad Khan}, \bibinfo{person}{Peican Zhu}, {and}
  \bibinfo{person}{Zhaoquan Gu}.} \bibinfo{year}{2022}\natexlab{}.
\newblock \showarticletitle{NormalAttack: Curvature-Aware Shape Deformation
  along Normals for Imperceptible Point Cloud Attack}.
\newblock \bibinfo{journal}{\emph{Security and Communication Networks}}
  (\bibinfo{year}{2022}).
\newblock


\bibitem[Wang et~al\mbox{.}(2019)]%
        {10.1145/3326362}
\bibfield{author}{\bibinfo{person}{Yue Wang}, \bibinfo{person}{Yongbin Sun},
  \bibinfo{person}{Ziwei Liu}, \bibinfo{person}{Sanjay~E. Sarma},
  \bibinfo{person}{Michael~M. Bronstein}, {and} \bibinfo{person}{Justin~M.
  Solomon}.} \bibinfo{year}{2019}\natexlab{}.
\newblock \showarticletitle{Dynamic Graph CNN for Learning on Point Clouds}.
\newblock \bibinfo{journal}{\emph{ACM Trans. Graph.}} (\bibinfo{year}{2019}).
\newblock


\bibitem[Wen et~al\mbox{.}(2024)]%
        {wen2024pointwavelet}
\bibfield{author}{\bibinfo{person}{Cheng Wen}, \bibinfo{person}{Jianzhi Long},
  \bibinfo{person}{Baosheng Yu}, {and} \bibinfo{person}{Dacheng Tao}.}
  \bibinfo{year}{2024}\natexlab{}.
\newblock \showarticletitle{PointWavelet: Learning in spectral domain for 3-D
  point cloud analysis}.
\newblock \bibinfo{journal}{\emph{IEEE Transactions on Neural Networks and
  Learning Systems}} (\bibinfo{year}{2024}).
\newblock


\bibitem[{Wen} et~al\mbox{.}(2020)]%
        {9294112}
\bibfield{author}{\bibinfo{person}{Y. {Wen}}, \bibinfo{person}{J. {Lin}},
  \bibinfo{person}{K. {Chen}}, \bibinfo{person}{C.~L.~P. {Chen}}, {and}
  \bibinfo{person}{K. {Jia}}.} \bibinfo{year}{2020}\natexlab{}.
\newblock \showarticletitle{Geometry-Aware Generation of Adversarial Point
  Clouds}.
\newblock \bibinfo{journal}{\emph{IEEE Transactions on Pattern Analysis and
  Machine Intelligence}} (\bibinfo{year}{2020}).
\newblock


\bibitem[Wu et~al\mbox{.}(2015)]%
        {wu20153d}
\bibfield{author}{\bibinfo{person}{Zhirong Wu}, \bibinfo{person}{Shuran Song},
  \bibinfo{person}{Aditya Khosla}, \bibinfo{person}{Fisher Yu},
  \bibinfo{person}{Linguang Zhang}, \bibinfo{person}{Xiaoou Tang}, {and}
  \bibinfo{person}{Jianxiong Xiao}.} \bibinfo{year}{2015}\natexlab{}.
\newblock \showarticletitle{3d shapenets: A deep representation for volumetric
  shapes}. In \bibinfo{booktitle}{\emph{CVPR}}.
\newblock


\bibitem[Xiang et~al\mbox{.}(2019)]%
        {xiang2019generating}
\bibfield{author}{\bibinfo{person}{Chong Xiang}, \bibinfo{person}{Charles~R
  Qi}, {and} \bibinfo{person}{Bo Li}.} \bibinfo{year}{2019}\natexlab{}.
\newblock \showarticletitle{Generating 3d adversarial point clouds}. In
  \bibinfo{booktitle}{\emph{CVPR}}.
\newblock


\bibitem[Xiang et~al\mbox{.}(2021)]%
        {xiang2021walk}
\bibfield{author}{\bibinfo{person}{Tiange Xiang}, \bibinfo{person}{Chaoyi
  Zhang}, \bibinfo{person}{Yang Song}, \bibinfo{person}{Jianhui Yu}, {and}
  \bibinfo{person}{Weidong Cai}.} \bibinfo{year}{2021}\natexlab{}.
\newblock \showarticletitle{Walk in the cloud: Learning curves for point clouds
  shape analysis}. In \bibinfo{booktitle}{\emph{CVPR}}.
\newblock


\bibitem[Xu and Zhou(2021)]%
        {xu2021deep}
\bibfield{author}{\bibinfo{person}{Zhiqin~John Xu} {and} \bibinfo{person}{Hanxu
  Zhou}.} \bibinfo{year}{2021}\natexlab{}.
\newblock \showarticletitle{Deep frequency principle towards understanding why
  deeper learning is faster}. In \bibinfo{booktitle}{\emph{AAAI}}.
\newblock


\bibitem[Yang et~al\mbox{.}(2019)]%
        {yang2019adversarial}
\bibfield{author}{\bibinfo{person}{Jiancheng Yang}, \bibinfo{person}{Qiang
  Zhang}, \bibinfo{person}{Rongyao Fang}, \bibinfo{person}{Bingbing Ni},
  \bibinfo{person}{Jinxian Liu}, {and} \bibinfo{person}{Qi Tian}.}
  \bibinfo{year}{2019}\natexlab{}.
\newblock \showarticletitle{Adversarial attack and defense on point sets}.
\newblock \bibinfo{journal}{\emph{arXiv preprint arXiv:1902.10899}}
  (\bibinfo{year}{2019}).
\newblock


\bibitem[Yang et~al\mbox{.}(2024)]%
        {yang2024hiding}
\bibfield{author}{\bibinfo{person}{Mingyu Yang}, \bibinfo{person}{Daizong Liu},
  \bibinfo{person}{Keke Tang}, \bibinfo{person}{Pan Zhou},
  \bibinfo{person}{Lixing Chen}, {and} \bibinfo{person}{Junyang Chen}.}
  \bibinfo{year}{2024}\natexlab{}.
\newblock \showarticletitle{Hiding Imperceptible Noise in Curvature-Aware
  Patches for 3D Point Cloud Attack}. In \bibinfo{booktitle}{\emph{ECCV}}.
\newblock


\bibitem[Zhang et~al\mbox{.}(2023a)]%
        {10.1145/3581783.3612018}
\bibfield{author}{\bibinfo{person}{Kui Zhang}, \bibinfo{person}{Hang Zhou},
  \bibinfo{person}{Jie Zhang}, \bibinfo{person}{Qidong Huang},
  \bibinfo{person}{Weiming Zhang}, {and} \bibinfo{person}{Nenghai Yu}.}
  \bibinfo{year}{2023}\natexlab{a}.
\newblock \showarticletitle{Ada3Diff: Defending against 3D Adversarial Point
  Clouds via Adaptive Diffusion}. In \bibinfo{booktitle}{\emph{MM}}.
\newblock


\bibitem[Zhang et~al\mbox{.}(2023b)]%
        {zhang2023ada3diff}
\bibfield{author}{\bibinfo{person}{Kaixuan Zhang}, \bibinfo{person}{Hang Zhou},
  \bibinfo{person}{Jie Zhang}, \bibinfo{person}{Qidong Huang},
  \bibinfo{person}{Weiming Zhang}, {and} \bibinfo{person}{Nenghai Yu}.}
  \bibinfo{year}{2023}\natexlab{b}.
\newblock \showarticletitle{Ada3diff: Defending against 3d adversarial point
  clouds via adaptive diffusion}. In \bibinfo{booktitle}{\emph{ACM MM}}.
  \bibinfo{pages}{8849--8859}.
\newblock


\bibitem[Zhang et~al\mbox{.}(2024)]%
        {zhang2024fourier}
\bibfield{author}{\bibinfo{person}{Liangqi Zhang}, \bibinfo{person}{Yihao Luo},
  \bibinfo{person}{Haibo Shen}, {and} \bibinfo{person}{Tianjiang Wang}.}
  \bibinfo{year}{2024}\natexlab{}.
\newblock \showarticletitle{A fourier perspective of feature extraction and
  adversarial robustness}. In \bibinfo{booktitle}{\emph{IJCAI}}.
\newblock


\bibitem[Zhou et~al\mbox{.}(2019)]%
        {zhou2019dup}
\bibfield{author}{\bibinfo{person}{Hang Zhou}, \bibinfo{person}{Kejiang Chen},
  \bibinfo{person}{Weiming Zhang}, \bibinfo{person}{Han Fang},
  \bibinfo{person}{Wenbo Zhou}, {and} \bibinfo{person}{Nenghai Yu}.}
  \bibinfo{year}{2019}\natexlab{}.
\newblock \showarticletitle{Dup-net: Denoiser and upsampler network for 3d
  adversarial point clouds defense}. In \bibinfo{booktitle}{\emph{CVPR}}.
\newblock


\end{thebibliography}


\appendix
\begin{algorithm}[htp]
	\caption{PWAVEP: Point Cloud Purification via Spectral Graph Wavelets}
	\label{alg:pwavep}
	\LinesNumbered
	\KwIn{Adversarial point cloud $\mathcal{P}' \in \mathbb{R}^{N \times 3}$, target classifier $f$, hyperparameters $K, \alpha, \beta, \gamma$}
	\KwOut{Purified point cloud $\mathcal{P}''$}

	\BlankLine

	\tcp{\textbf{Stage 1: Graph Wavelet Transform (GWT)}}
	Construct K-NN graph $\mathcal{G}$ and compute Laplacian $L$ from $\mathcal{P}'$\;
	Apply GWT to obtain wavelet coefficients $\{\psi_s\}$ and scaling coefficients $\{\delta\}$ (Eq.~\ref{eq:gwt_coeffs})\;

	\BlankLine

	\tcp{\textbf{Stage 2: Spectral-Spatial Analysis}}
	Compute composite loss $\mathcal{L}$ using Eq.~\ref{hyper-alpha}\;
	Compute wavelet gradient score $\nabla_{\psi_{s,i}}\mathcal{L}$ for each point $p_i$ at each scale $s$\;
	Compute Local Sparsity Score $S_{\text{LSS}}(p_i)$ for each point using Eq.~\ref{eq:lss}\;
	Compute hybrid saliency score $S_{\text{hy}}(p_i)$ for each point using Eq.~\ref{eq:hybrid-saliency}\;
	Rank points based on $S_{\text{hy}}$ and partition into high-risk set $\mathcal{P}_{\text{high}}'$ and mid-risk set $\mathcal{P}_{\text{mid}}'$ (Eq.~\ref{eq:high_risk_set_simple}-\ref{eq:mid_risk_set_simple})\;

	\BlankLine

	\tcp{\textbf{Stage 3: Hierarchical Purification}}
	Initialize modified wavelet coefficients $\psi' \gets \psi$\;
	\For{each point $p_i \in \mathcal{P}_{\text{mid}}'$}{
	Identify the most malicious wavelet band $s_i^*$ using Eq.~\ref{eq:highest_energy_band}\;
	Attenuate the coefficient $\psi'_{s_i^*, i}$ according to Eq.~\ref{eq:mid_risk_filtering_single}\;
	}
	Reconstruct temporary point cloud $\tilde{\mathcal{P}}$ via IGWT using $\psi'$ (Eq.~\ref{eq:igwt_reconstruction})\;
	Remove high-risk points to obtain the final purified cloud $\mathcal{P}''$ (Eq.~\ref{eq:high_risk_removal})\;

	\BlankLine

	\Return{$\mathcal{P}''$}
\end{algorithm}

\section{Chebyshev Approximation}
\label{app_sec:Chebyshev Approximation}

\begin{figure*}[tp]
	\centering
	\includegraphics[width=0.8\linewidth]{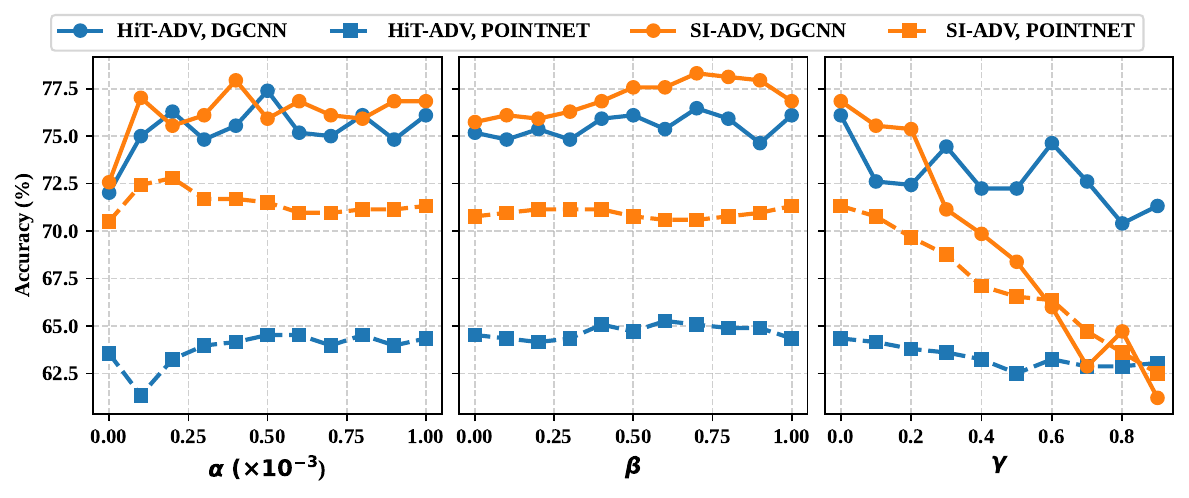}
	\caption{Hyper-parameter study on the ModelNet40 dataset. We analyze the impact of the feature stability weight $\alpha$ (left), the saliency balance weight $\beta$ (middle), and the filtering attenuation factor $\gamma$ (right) on the final defense accuracy. When adjusting one parameter, the others are set to their default values: $\alpha = 0.002, \beta = 1$, and $\gamma = 0$.}
	\label{fig:hyper-param}
\end{figure*}

\begin{figure*}[h]
	\centering
	\includegraphics[width=1\linewidth]{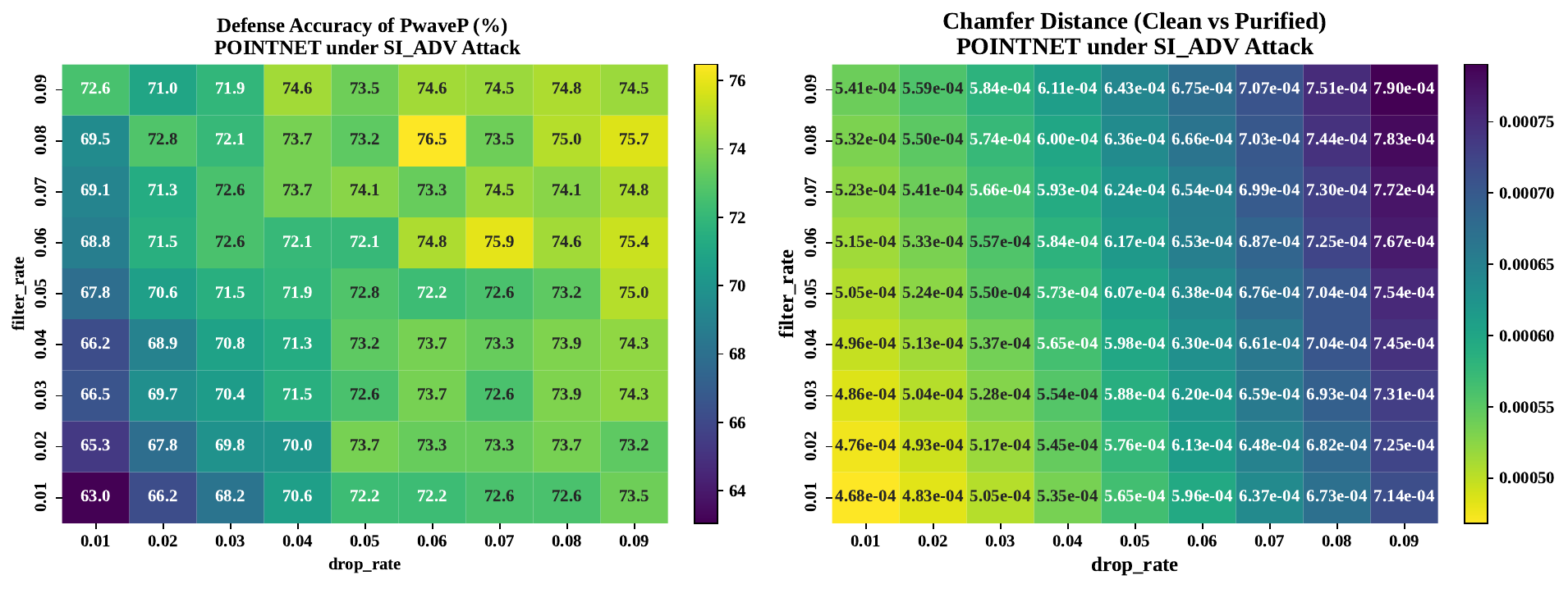}
	\caption{Left: Sensitivity analysis of defense accuracy (\%) with respect to the high-risk point removal rate (\texttt{drop\_rate}). Right: Impact of \texttt{drop\_rate} on geometric distortion, measured by Chamfer Distance (CD), where lower is better. Both evaluations use PointNet against the SI-ADV attack on the ModelNet40 dataset.}
	\label{fig:partition_accuracy}
\end{figure*}

\begin{figure*}[h]
	\centering
	\includegraphics[width=1\linewidth]{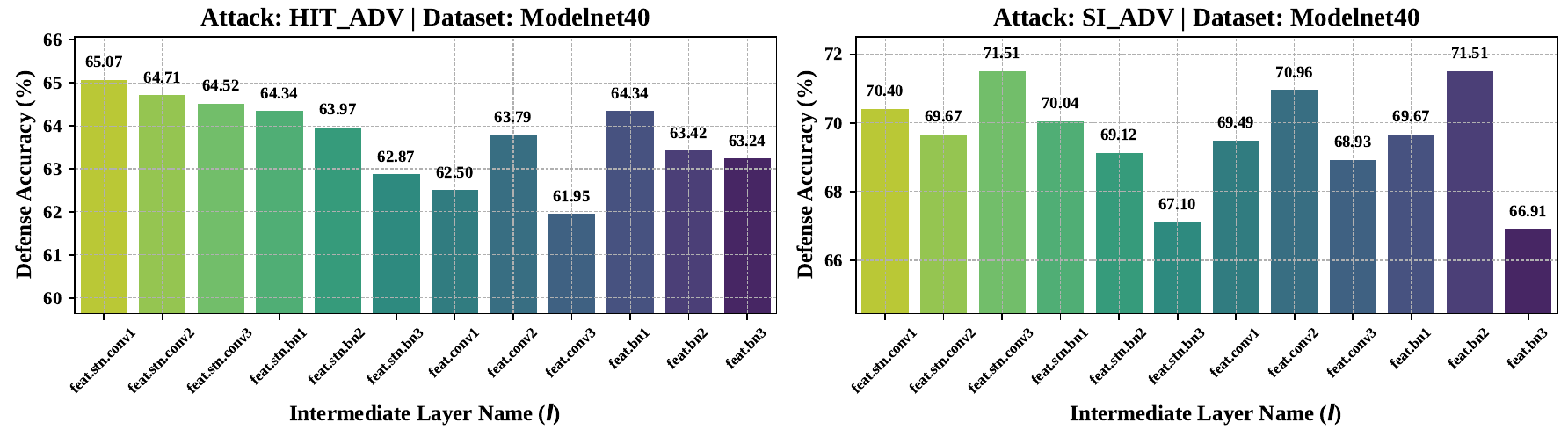}
	\caption{Evaluation of defense accuracy when using different intermediate layers of PointNet for the composite loss term. The analysis spans two attacks and two datasets. The results show no single layer is universally optimal; the best-performing layer is context-dependent.}
	\label{fig:layer_impact}
\end{figure*}
\begin{figure*}[t]
	\centering
	\includegraphics[width=0.8\linewidth]{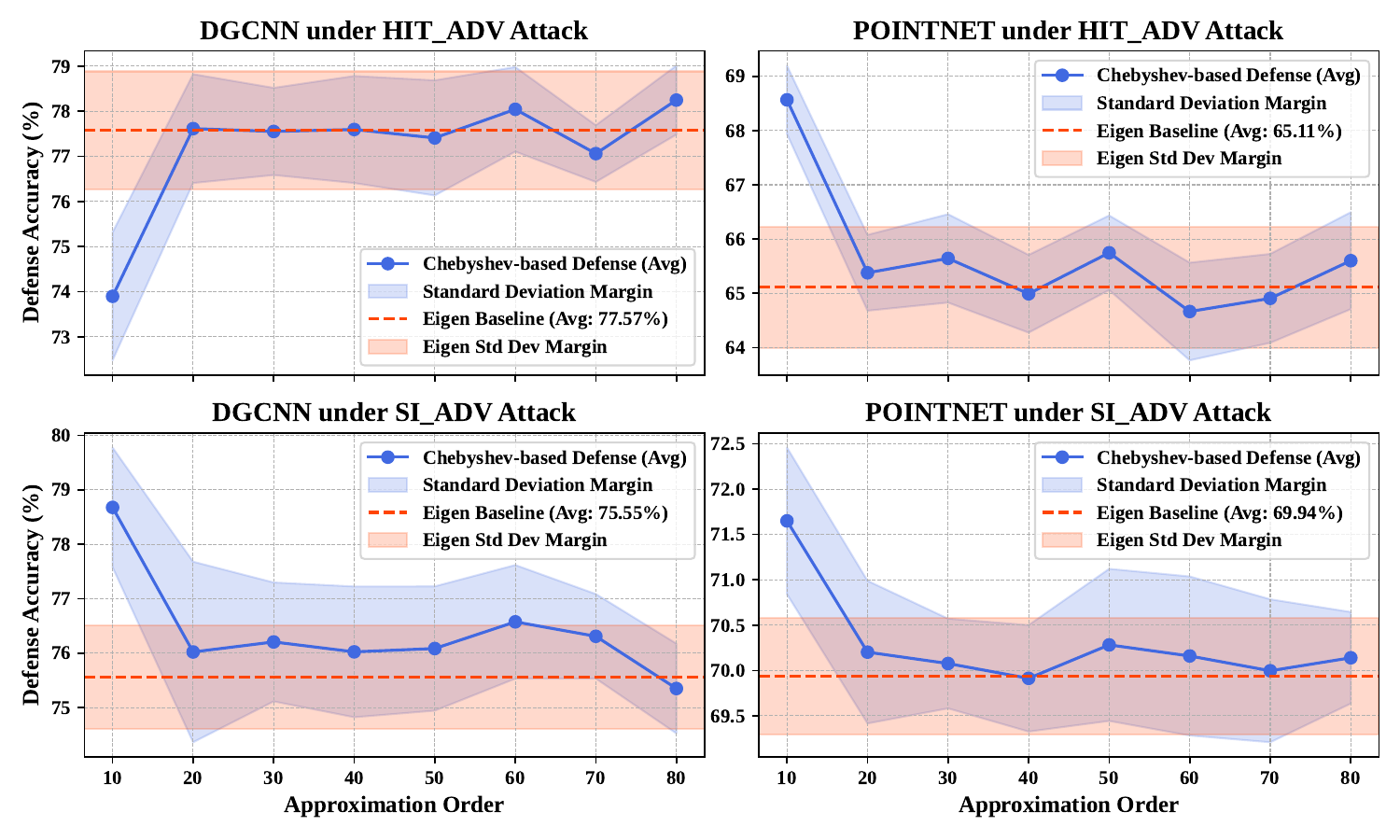}
	\caption{Defense accuracy of PWAVEP using exact eigen-decomposition versus its efficient Chebyshev approximation of varying orders. The evaluation is on ModelNet40 against HIT-ADV and SI-ADV attacks. Shaded areas represent standard deviation. Notably, the Chebyshev approximation can outperform the exact method, suggesting a potential implicit regularization effect.}
	\label{fig:cheb_vs_eigen}
\end{figure*}

Recall that in the Graph Wavelet Transform, there are two key components: the wavelet kernel at scale $s\in\{1,2,\dots,S\}$, defined as:
\begin{equation}
	T_s = g_s(L) = U g_s(\Lambda) U^\top,
\end{equation}
and the scaling function:
\begin{equation}
	T_\phi = g_\phi(L) = U g_\phi(\Lambda) U^\top.
\end{equation}
Note that Chebyshev Approximation requires the domain of the independent variable $x$ to lie within $[-1, 1]$ for the function $f(x)$ to be approximated. Therefore, in practice, we prefer using the normalized Laplacian matrix, defined as:
\begin{equation}
	\hat{L} = I - D^{-\frac{1}{2}} A D^{-\frac{1}{2}},
\end{equation}
where $A$ is the adjacency matrix and $D$ is the diagonal degree matrix, with $D_{ii} = \sum_j A_{i,j}$. The eigenvalues of the normalized Laplacian matrix $\hat{L}$ lie in $[0, 2]$, which conveniently aligns with the handling of the domain of $\Lambda$ in $g_s$ and $g_\phi$.
Chebyshev Approximation can be applied to both $T_s$ and $T_\phi$, where $s \in \{1, 2, \dots, S\}$. For simplicity, we denote:
\begin{equation}
	W = [g_\phi(\hat{L})^\top, g_1(\hat{L})^\top, g_2(\hat{L})^\top, \dots, g_S(\hat{L})^\top]^\top.
\end{equation}
According to \cite{HAMMOND2011129}, the approximations are given by:
\begin{equation}
	\tilde{g}_\phi(\hat{L}) = \frac{1}{2} c_{0,0} + \sum_{z=1}^Z c_{0,z} T_z(\hat{L}),
\end{equation}
\begin{equation}
	\tilde{g}_s(\hat{L}) = \frac{1}{2} c_{s,0} + \sum_{z=1}^Z c_{s,z} T_z(\hat{L}),
\end{equation}
where $\tilde{g}_\phi$ and $\tilde{g}_s$ are the approximations of $g_\phi$ and $g_s$, respectively; $Z$ is the number of terms in the truncated Chebyshev expansion; and $T_z(\hat{L})$ is the $z$-th Chebyshev polynomial, recursively defined as:
\begin{equation}
	T_z(\hat{L}) = 2 \left( \frac{2}{\lambda_{\max}} \hat{L} - I \right) T_{z-1}(\hat{L}) - T_{z-2}(\hat{L}),
\end{equation}
with $T_0(\hat{L}) = I$ and $T_1(\hat{L}) = \frac{2}{\lambda_{\max}} \hat{L} - I$. The coefficients $c_{0,z}$ and $c_{s,z}$ are the $z$-th Chebyshev coefficients, expressed as:
\begin{equation}
	c_{0,z} = \frac{2}{\pi} \int_{0}^{\pi} \cos(z \theta) \, g_\phi \left( \frac{\lambda_{\max} (\cos \theta + 1)}{2} \right) \, d\theta,
\end{equation}
\begin{equation}
	c_{s,z} = \frac{2}{\pi} \int_{0}^{\pi} \cos(z \theta) \, g_s \left( \frac{\lambda_{\max} (\cos \theta + 1)}{2} \right) \, d\theta.
\end{equation}
We denote the approximated transform operator as:
\begin{equation}
	\tilde{W} = [\tilde{g}_\phi(\hat{L})^\top, \tilde{g}_1(\hat{L})^\top, \tilde{g}_2(\hat{L})^\top, \ldots, \tilde{g}_S(\hat{L})^\top]^\top.
\end{equation}
When $g_\phi(\lambda)^2 + \sum_{s=1}^S g_s(\lambda)^2 = 1$ for all $\lambda \in \Lambda$, the transform satisfies the Parseval tight frame property, meaning the signal energy remains equal before and after the transform. In practice, this condition is often achieved by selecting specific kernel functions (such as those from the Meyer or Mexican hat families) and adjusting parameters accordingly. Otherwise, the Moore--Penrose pseudoinverse of the transform operator $\tilde{W}$ must be computed as:
\begin{equation}
	\tilde{W}^+ = (\tilde{W}^\top \tilde{W})^{-1} \tilde{W}^\top,
\end{equation}
and the inverse graph wavelet transform of the coefficients $c = \tilde{W} h$ for a signal $h \in \mathbb{R}^N$ can then be expressed as:
\begin{equation}
	\tilde{h} = \tilde{W}^+ c,
\end{equation}
where $\tilde{W} \in \mathbb{R}^{(S+1)N \times N}$, $\tilde{W}^+ \in \mathbb{R}^{  N \times (S+1)N}$, and $N$ is the number of nodes in the graph.

For a graph with $N$ nodes, the Chebyshev polynomial approximation of order $Z$ involves matrix-vector multiplications with the sparse Laplacian, resulting in a complexity of $O(Z \cdot E)$, where $E$ is the number of edges (often $O(N)$ in sparse graphs). Since $Z$ is typically small (e.g., 10--50) and independent of $N$, this yields a near-linear time complexity, making it far more efficient for large graphs compared to the $O(N^3)$ cost of eigen-decomposition.

\section{Hyper-Parameter Study}
We analyze the sensitivity of PWAVEP to its three main hyper-parameters to assess its robustness and ease of deployment. The study evaluates the feature stability weight $\alpha$, saliency balance weight $\beta$, and filtering attenuation factor $\gamma$. We use DGCNN and PointNet on the ModelNet40 dataset against HiT-ADV and SI-ADV attacks. Results are shown in Figure~\ref{fig:hyper-param}.

\section{Impact of Intermediate Layer Choice}\label{sec:Intermediate Layer Choice}
The choice of the intermediate layer $l$ for computing the feature stability loss is another hyper-parameter. We evaluate its impact on PointNet, whose modular architecture provides a clear test case.

As illustrated in Figure~\ref{fig:layer_impact}, our findings show that no single layer is universally optimal. The best choice is highly context-dependent, varying with both the attack type and the dataset. For instance, under HIT-ADV on ModelNet40, shallower layers (e.g., in the Spatial Transformer Network) perform well. In other scenarios, the choice is less critical. The key takeaway is that the layer $l$ should be treated as a tunable hyper-parameter. For optimal performance in a specific application, we recommend users perform a brief evaluation on a small validation set to identify the most effective layer for their model, threat landscape, and data distribution.

The results demonstrate that defense accuracy is stable across wide ranges of the feature stability weight ($\alpha$) and the saliency balance weight ($\beta$), confirming that PWAVEP is not highly sensitive to their precise settings and is practical to deploy. In stark contrast, the performance shows a decisive dependence on the attenuation factor $\gamma$. Accuracy is maximized at $\gamma=0$ (complete suppression) and consistently decreases as the suppression weakens (as $\gamma \to 1$). This finding provides powerful empirical validation for our core hypothesis: the high-frequency wavelet components identified by our saliency analysis are the primary carriers of adversarial noise, and their aggressive filtering is the optimal purification strategy.

\section{Impact of Partition Policy}
Figure~\ref{fig:partition_accuracy} illustrates the performance trade-off between defensive accuracy (color map) and geometric fidelity (Chamfer distance contours) as we vary the \texttt{drop\_rate} and \texttt{filter\_rate}. The analysis, conducted on ModelNet40 under SI-ADV attacks, reveals a distinct optimal region where a low \texttt{drop\_rate} ($\approx 1\%$) and a high \texttt{filter\_rate} ($\approx 9\%$) maximize accuracy while preserving geometric integrity. Crucially, an overly aggressive \texttt{drop\_rate} proves counterproductive, creating a "lose-lose" scenario where both accuracy and fidelity degrade. Developing an adaptive policy to find the optimal partition automatically is a valuable direction for future work.

\section{Eigen-Decomposition v.s. Chebyshev Approximation}
We use Chebyshev polynomial expansion to efficiently approximate the graph wavelet transform, avoiding the prohibitive cost of direct eigen-decomposition. While motivated by efficiency, this approximation reveals a non-trivial impact on defense efficacy, as shown in Figure~\ref{fig:cheb_vs_eigen}.

Counter-intuitively, the approximated version does not simply trail the performance of the exact eigen-decomposition baseline. In several cases, the Chebyshev approximation provides consistently better accuracy. A plausible hypothesis for this phenomenon is \textit{implicit regularization}. An exact filter perfectly mirrors the spectral properties of the adversarial graph, potentially "overfitting" to the attacker's malicious patterns. In contrast, the Chebyshev approximation, being an imperfect representation, introduces minor deviations that may smooth the filter response, preventing over-sensitivity to high-frequency adversarial noise and thus inadvertently enhancing robustness.

While this observation is compelling, a full theoretical characterization is beyond our current scope. We therefore interpret this effect with caution and do not claim it as a designed feature. Based on the empirical results, we recommend a Chebyshev order between 30 and 50, which offers a stable and effective balance of performance and computational cost. Formal analysis of this implicit regularization is a promising direction for future work.

\end{document}